\documentclass{article}

    \PassOptionsToPackage{numbers}{natbib}
 \usepackage[preprint]{neurips_2026}

\usepackage{subcaption}
\usepackage[table]{xcolor}
\usepackage{colortbl}
\usepackage{transparent} 
\usepackage{multirow}
\usepackage{pifont}
\usepackage[most]{tcolorbox}
\usepackage{bbm}
\setkeys{Gin}{draft=false}
\usepackage{algorithm}
\usepackage{algpseudocode}
\usepackage{microtype}    

\definecolor{boxgray}{RGB}{245,245,245}
\newtcolorbox{mybox}[1]{ colback=boxgray, colframe=black, boxrule=0.8pt, arc=3pt, boxsep=4pt, left=6pt, right=6pt, top=6pt, bottom=6pt, title=#1, fonttitle=\bfseries, boxed title style={size=small,boxrule=0.8pt}, coltitle=white, colbacktitle=black, before skip=10pt, after skip=10pt }

\usepackage[utf8]{inputenc} 
\usepackage[T1]{fontenc}    
\usepackage{hyperref}       
\usepackage{url}            
\usepackage{booktabs}       
\usepackage{amsfonts}       
\usepackage{nicefrac}       
\usepackage{microtype}      
\usepackage{xcolor}         

\title{DirectSpeech2LLM: A Simple End-to-End Framework to Mitigate Prompt Overfitting in Speech-LLMs}

\author{%
  Hemant Yadav\thanks{indicates the corresponding author.} \\
  IIIT Delhi, India\\
  \texttt{hemantya@iiitd.ac.in} \\
  \And
  Sunayana Sitaram\\
  Microsoft Research, India  \\
  \texttt{sunayana.sitaram@microsoft.com} \\
  \And
  Roger Zimmermann\\
  National University of Singapore \\
  \texttt{dcsrz@nus.edu.sg} \\
  \And
  Rajiv Ratn Shah \\
  IIIT Delhi, India\\
  \texttt{rajivratn@iiitd.ac.in} \\
}

\begin{document}

\maketitle

\begin{abstract}
  Speech-LLMs often exhibit prompt overfitting, where models solely trained on automatic speech recognition (ASR) instruction fail to generalize to new instructions such as speech translation and continue to behave primarily as ASR system. We propose \textbf{DirectSpeech2LLM}, a simple end-to-end framework that preserves the instruction-following ability of the LLM on unseen tasks when conditioned on speech. It computes distance-based CTC loss over the frozen LLM embedding matrix and uses greedy CTC labels to derive geometrically and temporally aligned speech embeddings respectively as an input to the LLM. Trained solely on 960 hours of LibriSpeech ASR data, DirectSpeech2LLM outperforms the cascaded system on ASR (seen task) and generalizes zero-shot to speech translation and emotion recognition (two unseen tasks), closely matching the cascaded system upper bound on these two new instructions despite seeing neither during training. 
  We also find that geometric alignment strength plays a smaller role than previously assumed, as our modified CTC loss is shown to provide sufficient implicit geometric grounding without requiring an explicit regression loss.
  Results are consistent across two LLM families and scale with both more training data and model capacity.  

\end{abstract}

\begingroup
\renewcommand{\thefootnote}{}
\footnotetext{Code available at: \url{https://github.com/raotnameh/DirectSpeech2LLM}}
\endgroup

\section{Introduction}
\label{sec:intro}

\begin{figure}[t]
\centering
\includegraphics[width=\textwidth]{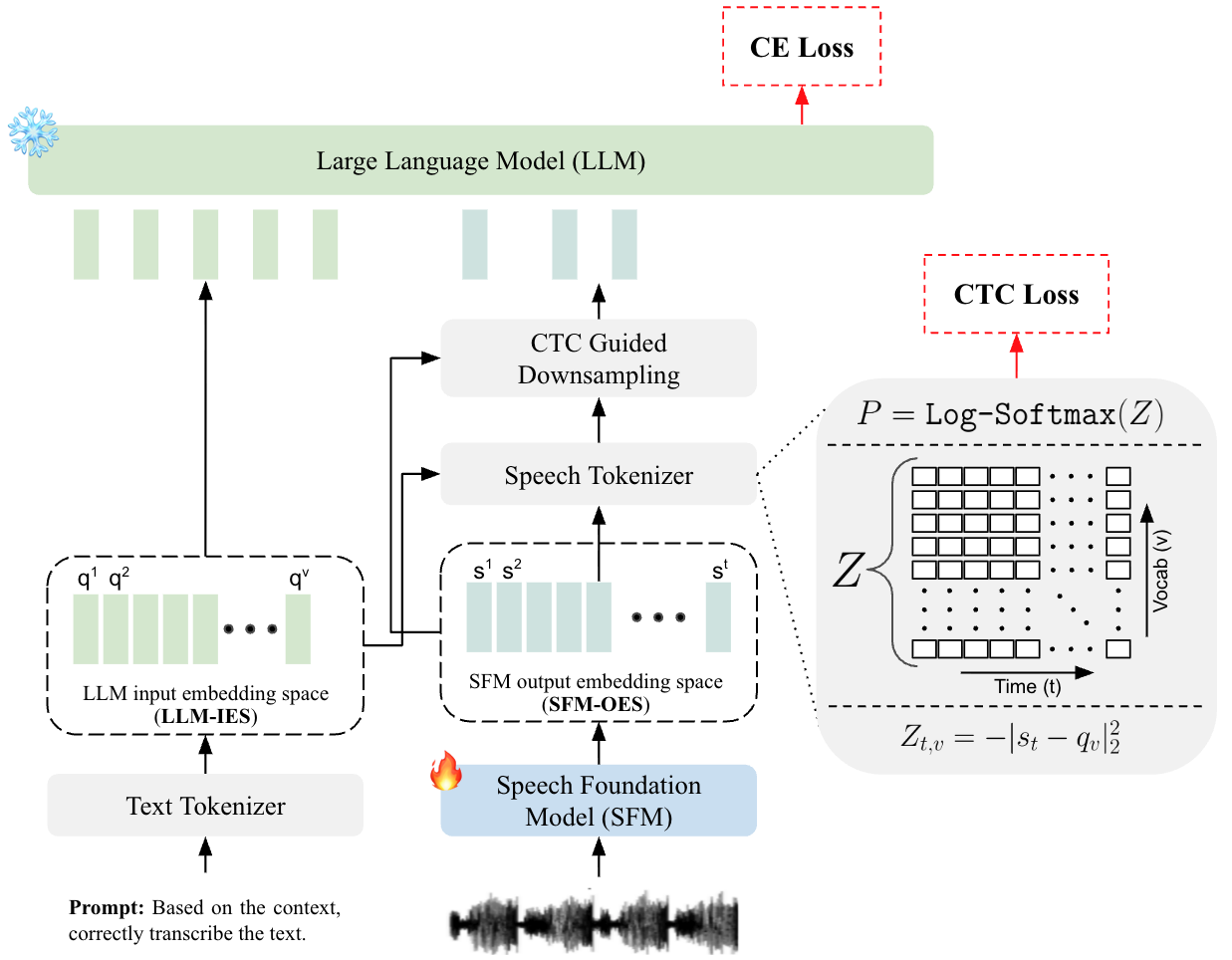}
\caption{Overview of \textbf{DirectSpeech2LLM}. Given an input speech signal, the SFM produces frame representations $s_t \in \mathbb{R}^d$, where $d$ matches the dimensionality of the LLM input embeddings $Q$. Logits are computed as negative squared distances ($Z_{t,v}$) followed by log-softmax and CTC loss ($\mathcal{L}_{\text{ctc}}$)  is applied. Next, the resulting greedy CTC predicted labels are then used to downsample SFM output embeddings via mean pooling. The LLM is fed with the resulting updated sequence embeddings from SFM and is trained with cross-entropy loss ($\mathcal{L}_{\text{llm}}$). The reader should keep in mind that the LLM-IES serves only as a geometric anchor for the SFM-OES, without being replaced or discretized with the LLM input embeddings. Commitment loss is not shown in the figure for clarity.}
\label{figure:proposed_method}
\end{figure}

Large language models (LLMs) have shown a remarkable ability to learn rich linguistic information from raw text, enabling strong generalization across language understanding, generation, and reasoning tasks \cite{brown2020language, touvron2023llama, openai2023gpt4}. 
Speech, in contrast, is a continuous acoustic signal that conveys the same underlying linguistic information that LLMs already represent effectively in discrete textual form.
The ability of a pretrained LLM (the “brain”) to process auditory inputs (the “ears”) while preserving its original generalization ability is referred to as a desirable Speech-LLM \cite{Peng_2026_survey_speech_llm}. 
Speech also encodes other information such as speaker identity, prosody, and noise \cite{yadav2024jooci}.
However, current large audio language models are shown to behave largely as transcription systems, relying on lexical semantics and underutilizing acoustic cues \cite{chen2025audiollmsreallylisten}. 
This work focuses on improving the linguistic modeling of Speech-LLMs to new tasks not seen during training and leave improving non-linguistic modeling for future.

Enabling LLMs to process linguistic content present in speech requires transforming speech into embeddings aligned with the LLM input embedding space (IES) along two dimensions: geometric and temporal \cite{Deng2024Wav2PromptES}. 
Geometric alignment ensures that speech derived embeddings are consistent with the geometry of the trained LLM-IES by minimizing a distance metric.
Temporal alignment ensures that the embedding sequence exhibits statistical properties consistent with those observed during LLM training. 

Existing Speech-LLMs connect a pretrained speech foundation model (SFM) to an LLM using a projection module\footnote{for simplicity, from now on we assume the projection module is also part of SFM.}, implemented using a linear layer \cite{gao2022wavpromptfewshotspokenlanguage, wu2023decoderctcdownsample1}, non-linear layers \cite{ma2024embarrassinglysimpleapproachllmslamasr}, CNN layers \cite{hu2024wavllm}, Q-former \cite{tang2023salmonn} or a combination of them. 
While effective on seen tasks, current models often fail to generalize to unseen tasks \cite{fan2024alignformer} or prompt overfitting\footnote{The terms \textit{prompt overfitting} and \textit{failure to generalize to unseen tasks} are used interchangeably to describe the same phenomenon. Both refer to the loss of the original instruction-following ability of the underlying LLM in Speech-LLMs}.
For example, a Speech-LLM model trained solely on ASR task continues to behave as an ASR system even when prompted differently to perform translation or question answering \cite{Peng_2026_survey_speech_llm, cuervo2026closinggaptextspeech}.

We hypothesize that this lack of generalization arises because of two factors. 
First, the SFM output embedding space (OES) is not constrained geometrically with the LLM-IES, so embeddings can occupy arbitrary regions in $\mathbb{R}^d$ during training, degrading the LLM's generalization ability.
Prior work has shown that LLMs do tolerate small perturbations in their input embeddings, but performance degrades sharply beyond a threshold \cite{mohapatra2026speechmapper}. 
Since the LLM is frozen, only the SFM outputs can be constrained to lie close to the LLM-IES, otherwise the system reduces to a task-specific pipeline similar to Whisper \cite{radford2023robustwhisper}.

Second, most approaches rely on fixed temporal downsampling \cite{zhang2023speechgpt, pan2023cosmicspeechintsrtcutiontunign, chu2023qwenaudioadvancinguniversalaudioqwen1}, which is inherently misaligned with the highly variable temporal dynamics of continuous speech.
Because phoneme durations vary across speakers, speaking rates, and linguistic contexts, fixed downsampling produces inconsistent sequence boundaries for identical linguistic units causing overfitting.
Consequently, the embedding sequence length for a given token becomes an artifact of speaking rate rather than semantic content.
Given the LLM is fixed, such temporal mismatch shifts the input distribution away from that seen during LLM's training, impacting its generalization ability \cite{koh2021wilds}.

In this work, we propose \textbf{DirectSpeech2LLM}, a simple end-to-end Speech-LLM framework that jointly enforces geometric and temporal alignment. Figure~\ref{figure:proposed_method} illustrates the overall framework. To save compute time DirectSpeech2LLM is trained in two stages to save on compute time. 
In Stage~1, the SFM is trained using a modified CTC objective augmented with vector quantization (VQ) to enforce alignment along the two dimensions of LLM. In Stage~2 (end-to-end), we add the LLM cross-entropy loss to the Stage~1 objectives to further improve performance.

AlignFormer enforces temporal alignment by dynamically downsampling the speech embeddings using Connectionist Temporal Classification (CTC) to match the LLM's input temporal distribution \cite{fan2024alignformer}. However, they do not enforce geometric constraint between the SFM-OES and LLM-IES. A similar work, Wav2Prompt \cite{Deng2024Wav2PromptES} uses continuous integrate-and-fire (CIF) module to align temporally and regression loss to map speech embeddings to equivalent LLM tokens but suffers from a train/inference mismatch due to length dependent weight scaling and the performance is very sensitive to geometric alignment strength on unseen tasks, unlike our method.

Empirically, DirectSpeech2LLM achieves strong performance on ASR (seen task) and generalizes zero-shot to speech translation and emotion recognition (two unseen tasks or two new instructions).
DirectSpeech2LLM also demonstrates robustness to out-of-domain setting on the ASR task, similar to CTC trained methods, which is shown to be a limitation of simple adapter based methods \cite{kumar2025performancerobustness1, oh2026seamrobustness2}.

\vspace{0.2cm} 
\noindent Our main contributions are as follows.
\begin{enumerate}
    \item 
    We propose DirectSpeech2LLM, a simple end-to-end Speech-LLM framework that preserves LLM's generalization ability to new instructions without relying on (i) large amount of multitask instruction tuning dataset or (ii) costly multiple forward passes to LLM.

    \item Our method treats CTC as a primary objective rather than an auxiliary loss, giving the SFM three independently usable operating modes at test time: (i) standalone CTC-based ASR, (ii) CTC $+$ n-gram LM ASR, or (iii) full end-to-end Speech-LLM. 
    Unlike most prior approaches that tightly couple the SFM and LLM into a single pipeline, SFM output embeddings here are fed \textit{directly} to the LLM without any projection module, discretization, or separate adapter -- hence the name \textbf{DirectSpeech2LLM}.

    
\end{enumerate}

The rest of the paper is organized as follows: Section \ref{sec:background} reviews related work, Section \ref{sec:methods} describes the proposed method, Section \ref{sec:expdetails} details the experimental setup, Section \ref{sec:results} presents results and ablations, and Section \ref{sec:conclusion} concludes with future directions and limitations.

\section{Related works}
\label{sec:background}
Recent advancements in Speech-LLMs have explored various strategies to align SFM with LLM. One common direction focuses on solving the modality gap through behavior alignment. BLSP \cite{wang2023blsp} trains Speech-LLMs to mimic LLM responses to repeat task based prompts. While this improves zero-shot capabilities, performance lags behind instruction-tuned models. SALAD \cite{cuervo2026closinggaptextspeech} extends this work by combining cross-modal distillation with targeted synthetic data to improve alignment. But behavior alignment methods require two LLM forward passes during training, roughly doubling compute, compared to one forward pass in instruction tuned methods. Other approaches rely on computationally expensive large-scale synthetic or data-driven pretraining to preserve LLM generalization capabilities \cite{zeng2024scalingspeechtextpretrainingsyntheticmoredata, tang2023salmonn, hu2024wavllm, zhang2025soundwave}. Behavior alignment is applied on the LLM-OES while our method is aligning the LLM-IES. Therefore, these two are complementary and can be applied together.

Another direction, focuses on temporal alignment using CTC. In the context of Speech-LLMs AlignFormer \cite{fan2024alignformer} utilizes CTC alignments as a temporal guide to form dynamic windows for a Q-Former adapter. A similar work is Soundwave \cite{zhang2025soundwave} which uses simple projection layers instead of Q-former. LegoSLM \cite{ma2025legoslm} bypass adapters completely by utilizing CTC posterior distributions and finetune the LLM. It computes a weighted sum of the LLM’s existing text embeddings based on CTC posteriors to create pseudo-speech embeddings.
While TASU \cite{peng2025tasu} uses simulated pseudo CTC erroneous text during training to learn the adapter without using paired speech data.

Prior speech translation work \cite{tsiamas2024pushingspeechtransaltionctclaignnmentoptimaltransport2, le2023prespeechtransaltionctclaignnmentoptimaltransport2} separates temporal alignment (CTC) and distribution-level geometric alignment (Optimal Transport). DirectSpeech2LLM unifies both by computing CTC logits as negative squared distances to frozen LLM embeddings as fixed codebooks, jointly achieving dynamic temporal compression and geometric grounding.

Unlike Wav2Prompt \cite{Deng2024Wav2PromptES},  which uses CIF followed by regression, DirectSpeech2LLM uses modified CTC to compute logits as negative squared distances to the frozen LLM embedding matrix. As CTC is the canonical ASR framework, DirectSpeech2LLM provides a conceptually simpler and robust, single objective solution that unifies temporal and geometric alignment without extra hyper parameters or mismatch between train/inference. Wav2Prompt also collapses if regression loss (geometric) is removed compared to our method where commitment loss (geometric) is optional as CTC implicit has geometric constraint, making our method simpler. Lastly the geometric loss used in Wav2Prompt does not take into account negative LLM token embeddings, whereas our distance-based logits i.e., move the positive embedding closer while negative embeddings are pushed farther apart. This could be the reason for our method's less sensitivity to the regression loss 

Different to AlignFormer \cite{fan2024alignformer}, which uses CTC to handle temporal alignment but leaves the SFM-OES geometrically unconstrained. DirectSpeech2LLM enforces geometric grounding using distance-based logits (implicit) and a commitment constraint (explicit). Which is shown to prevent prompt overfitting without relying on input ordering tricks as in Alignformer. Again, AlignFormer input ordering trick is complementary to our method both can be applied together.

\section{Method}
\label{sec:methods}

\subsection{Preliminaries}
Please refer to Section \ref{sec:appendix_background} for preliminaries.

\subsection{Problem formulation}

Let $X = (x_1, \dots, x_M)$ denote an input acoustic sequence and 
$Y = (y_1, \dots, y_U)$ its corresponding ground truth labels, where 
$y_u \in \mathcal{V}_{\text{ctc}} \subseteq \mathcal{V}_{\text{llm}}$.
We assume a trained LLM ($\mathcal{M}_\theta$) with frozen parameters $\theta$ and embedding matrix $Q \in \mathbb{R}^{|\mathcal{V}_{\text{llm}}| \times d_{\text{llm}}}$. where each row $q_j$ uniquely corresponds to one of the LLM's vocabulary embedding. The SFM produces frame level embeddings with the same dimension as $\mathcal{Q}$.
\[
E_\phi : X \mapsto S = (s_1, \dots, s_T), 
\quad s_t \in \mathbb{R}^{d_{\text{llm}}},
\]
Our objective is to learn $\phi$ such that $S$ is aligned with LLM-IES i.e., each $s_t$ lies closer to its equivalent in $Q$ in both temporal and geometric dimensions.

\subsection{Speech tokenizer}
\label{sec:ctcloss}

Speech Tokenizer computes frame level logits given (i) SFM output embeddings and (ii) LLM input embedding matrix $\mathcal{Q}$ using negative squared Euclidean distance similar to \cite{van2017neuralvqvae}.
\begin{equation}
Z_{t,v} = - | s_t - q_v |_2^2.
\end{equation}
Expanding,
\begin{equation}
Z_{t,v}
= - \left(
\lVert s_t \rVert _2^2 
- 2 s_t^\top q_v 
+ \lVert q_v \rVert _2^2
\right).
\end{equation}
Since $\lVert s_t \rVert _2^2$ does not depend on $v$, this formulation is equivalent to a linear projection with fixed weights $Q$. This is equivalent to selecting the nearest token embedding in the LLM input embedding space. 

The LLM has a large vocabulary $\mathcal{V}_{\text{llm}}$ in thousands if not hundreds of thousands. Directly using this vocabulary for CTC loss is suboptimal. CTC training is more stable and sample efficient when prediction units correspond to smaller, frequently occurring acoustic segments.

To obtain acoustically suitable prediction units, we instead construct a dedicated CTC vocabulary $\mathcal{V}_{\text{ctc}}$ using byte-pair encoding (BPE) trained on the normalized LibriSpeech transcripts\footnote{\url{https://openslr.trmal.net/resources/11/librispeech-lm-norm.txt.gz}}.
We use a vocabulary size of 1000, which is a widely adopted starting point in end-to-end ASR systems, particularly for CTC and seq2seq baselines on the LibriSpeech dataset, as it provides a favorable trade-off between sequence length and acoustic consistency\footnote{\url{https://speechbrain.readthedocs.io/en/stable/tutorials/tasks/speech-recognition-from-scratch.html}}.
In the next step, we discard tokens which do not appear in the Qwen-3 LLM tokenizer vocabulary, resulting in 984 valid tokens for computing CTC loss (see Algorithm \ref{algorithm:ctc_vocab} in Section \ref{sec:appendix_algorithms} for pseudo code). For experiments with Phi-3, we initialize BPE with 2000 merge operations to obtain a final retained subset comparable in size to that derived from Qwen-3. The reader should keep in mind that the LLM still generate/output tokens over its original full vocabulary space.

During training, frame level logits are computed over the full LLM embedding matrix $Q$ followed by log-softmax, meaning the negatives in the softmax function are consisting of the full LLM vocabulary. This is different from Wav2Prompt.
Our earlier experiments showed that it resulted in better geometric alignment during stage~1. The final CTC loss is calculated only over $\mathcal{V}_{\text{ctc}}$ vocabulary indices. We obtain: $ P_{\text{ctc}} = \mathrm{logsoftmax}(Z[ \mathcal{V}_{\text{ctc}}]).$ The CTC objective is then:
\begin{equation}
\mathcal{L}_{\text{ctc}}
= - \log 
\sum_{\pi \in \mathcal{B}^{-1}(Y)}
\prod_{t=1}^{T}
P_{\text{ctc}}(\pi_t \mid t),
\end{equation}
where $\mathcal{B}$ denotes the CTC collapse function. 
The modified CTC loss enforces explicit temporal alignment and implicit geometric alignment between the SFM-OES and the LLM-IES.

\subsection{CTC guided downsampling}

For each output frame of SFM, we compute a greedy label assignment:
\begin{equation}
\label{eq:ctc argmax}
\hat{I}_t = \arg\min_j | s_t - q_j |_2^2.
\end{equation}
Next, we partition the output embeddings into consecutive runs of identical predictions. Let $\mathcal{T}_i$ denote the set of indices belonging to the $i$-th run. Then each segment embedding is obtained by mean pooling the corresponding identical labels.
\[
s_i^{\mathrm{seg}} =
\frac{1}{|\mathcal{T}_i|}
\sum_{t \in \mathcal{T}_i} s_t, \qquad i = 1,\dots,U' .
\]
Collecting all segment embeddings yields $S_{\mathrm{seg}} \in \mathbb{R}^{U' \times d_{\mathrm{llm}}}$, where $U'$ is the number of runs. 

Lastly we global mean pool all the blank segments into 1 embedding. This step compresses the original SFM output embedding sequence of length $T$ into 1 blank embedding, followed by remaining non-blank embeddings. The position of blank embedding is always at the start and the equivalent token in $Q$ is underscore ( $\_$ ). 

Exploring learnt embeddings for blank token and its ability to capture paralinguistic information of the speech utterance is out of scope of this work and is left for future work.

\subsection{Commitment loss}
Using the greedy CTC labels from Equation~\ref{eq:ctc argmax}, we explicitly constrain SFM outputs to be close to the pretrained LLM-IES using a commitment loss:
\begin{equation}
\mathcal{L}_{\text{commit}}
=
\frac{1}{\sqrt{d_{\text{llm}}}}
\sum_{t=1}^{T}
M_t
| s_t - q_{\hat{I}_t} |_2^2.
\end{equation}
Where $M_t = \mathbbm{1}[\hat{I}_t \neq b]$ is the mask of valid indices excluding blank token $b$. Commitment loss is \textit{not} applied to the blank token indices because it does not encode linguistic information of the input utterance that has an equivalent in $Q$. Blank embedding is similar to Prefix-Tuning in LLM \cite{li2021prefixtuningoptimizingcontinuousprompts}, that is it is same for all the samples in training.

The reader should keep in mind that the modified CTC loss (in Section \ref{sec:ctcloss}) itself encourages the SFM output to lie closer to the LLM-IES geometrically, not as directly as using a commitment loss.

\subsection{LLM}

The segmented embeddings ($S_{\mathrm{seg}}$) are concatenated with prompt embeddings \(E_p\) (e.g., ``Prompt: Correctly transcribe the following text:''). The resulting interleaved sequence is fed to the LLM and is trained using a standard auto regressive cross-entropy loss over the target text sequence \(Y = (y_1,\dots,y_U)\).
\[
\mathcal{L}_{\text{llm}}
=
-
\sum_{u=1}^{U}
\log p_\theta(y_u \mid X = [E_p=Q(Prompt), S_{\text{seg}}], y_{<u}).
\]

\subsection{Overall training objective}

The final loss is:
\begin{equation}
\mathcal{L}
=
\mathcal{L}_{\text{ctc}}
+
\mathcal{L}_{\text{commit}}
+
\lambda_{\text{llm}} \mathcal{L}_{\text{llm}}.
\end{equation}

The commitment loss is clamped to never exceed CTC loss because its targets are derived from greedy CTC predictions, meaning commitment loss is logically downstream of CTC. If CTC predictions are poor, the greedy labels are noisy, and the commitment loss would be pushing SFM embeddings toward incorrect geometric anchors in the LLM-IES. 

Only the SFM parameters, $\phi$, are optimized and the LLM parameters are fixed. Training proceeds in two stages. In stage~1 we set $(\lambda_{\text{llm}} = 0)$. In the stage~2 (end-to-end) the LLM loss is enabled, allowing contextual feedback from the frozen LLM to further improve performance. 
For full training pseudocode see Algorithm~\ref{algorithm:1} in Section \ref{sec:appendix_algorithms}.

\section{Experimental details}
\label{sec:expdetails}

\subsection{Training data}

All models are trained using only the English ASR data (single task training setup) with SpecAugment \cite{Park2019SpecAugmentAS}.

\noindent\textbf{LibriSpeech: }
We train on the full 960-hour LibriSpeech corpus \cite{panayotov2015librispeech}. This is referred as 1k data.

\noindent\textbf{Common Voice: }
We additionally experiment with training on combined 1800 hours of the English subset of Mozilla Common Voice (version 24) \cite{ardila2020common} and 960 hours LibriSpeech. 
We use the official train split and no additional filtering or data cleaning is applied. This is referred as 2.8k data.

\subsection{Model configuration}

\textbf{LLM Backbone: }We use \texttt{Qwen3-4B}\footnote{\url{https://huggingface.co/Qwen/Qwen3-4B}} and \texttt{Phi3.1-3.8B}\footnote{\url{https://huggingface.co/microsoft/Phi-3-mini-128k-instruct}} from HuggingFace as the language model backbone. 
All parameters of the LLM, including its token embedding matrix $Q$, remain frozen during training.

\noindent\textbf{SFM backbone: }It is initialized from the ASR finetuned \texttt{Data2Vec-Large} \cite{baevski2022data2vec} checkpoint having 24 transformer layers ($\approx$300 million parameters). The original linear CTC head is removed and reinitialized to match the LLM input embedding dimension and is retrained with the modified CTC loss during stage 1.

\subsection{Training procedure}
Training proceeds in two stages. 
Stage~1 is computationally cheap to temporally align SFM-OES and act as a good starting point for stage~2 training to save on compute.
All training runs use only English ASR data, no other task specific supervised finetuning is done in any way, unless mentioned otherwise. All models are trained for a fixed number of optimization steps, rather than until convergence. Specifically, Stage 1 is trained for 50k steps, while Stage 2 is trained for 20k steps, unless mentioned otherwise. For more training details, please refer to Section \ref{sec:appendix_tp}.

\subsection{Evaluation procedure}
During training both the blank embeddings and non blank embedding were fed to the LLM (WB setup). During inference for the ASR task, we follow the same setup. But on unseen tasks, we find that not using blank tokens (NB setup) consistently gives better results. Therefore we do not use blank token when evaluating on unseen tasks (NB setup). We use different task prompts for each task as shown in Figure \ref{fig:taskprompt} in the Section \ref{sec:appendix_taskpromptsused}. We use mms text normalization\footnote{\url{https://github.com/facebookresearch/fairseq/blob/main/examples/mms/data_prep/text_normalization.py}} to normalize the LLM output and ground truth.

\noindent\textbf{Automatic Speech Recognition (ASR): }
We evaluate in-domain ASR performance on the LibriSpeech test-clean and test-other splits and out-of-domain (OOD) using VoxPopuli (VP) test set \cite{wang-etal-2021-voxpopuli}. 
The number of LLM output words are clamped at $\max(\text{CTC} + 5,\; \lfloor 1.1 \times \text{CTC} \rfloor)$.
Performance is measured using Word Error Rate (WER) computed with the \texttt{jiwer} toolkit.

\noindent\textbf{Speech Translation (ST): }
Zero-shot speech translation is evaluated on CoVoST2 \cite{wang2020covost2} dataset for English-German (en-de) and English-Chinese (en-zh) split.  No translation data is used during training. 
The LLM output characters are clamped at $\lfloor 2 \times \text{CTC} \rfloor$
Performance is measured using BLEU computed with the \texttt{SacreBLEU} toolkit\footnote{\texttt{SacreBLEU (v2.3.1): 
nrefs:1|case:mixed|eff:no\allowbreak|tok:13a,zh|smooth:exp}}.

\noindent\textbf{Spoken Question Answering (SQA): }
We evaluate zero-shot spoken question answering using the IEMOCAP emotion recognition (ER) dataset \cite{busso2008iemocap} with the standard 4-class (angry, happy, neutral, and sad) setup using set 5. 
Performance is reported using classification accuracy. This task is primarily to study the instruction following rate (IFR) of the model, that is, can the model follow new instructions effectively different from what seen during training?

\section{Results}
\label{sec:results}

The closest prior works to DirectSpeech2LLM are LegoSLM, TASU, and AlignFormer. LegoSLM uses CTC posterior probabilities to construct weighted LLM input embeddings and finetunes the LLM on paired speech-text data (ASR and en–de translation). TASU, in contrast, simulates CTC erroneous text to train the adapter using text-only data (ASR and en–zh translation). AlignFormer freezes the LLM and jointly trains the SFM with a dynamic Q-Former adapter using only ASR paired speech text data. 
Wav2Prompt is related but not directly comparable due to its smaller SFM and an older family of LLM backbone, we only compare its regression loss sensitivity to unseen tasks in Section~\ref{sec:appendix_comparsionwithwav2prompt}.

\renewcommand{\arraystretch}{1.0}
\begin{table*}[t]
\centering
\caption{
Models marked with \textit{SFT} uses supervised finetuning for that task.
DirectSpeech2LLM shows near perfect instruction following on two unseen tasks, ST and ER as performance approaches to the upper bound of cascade system showing LLM's generalization ability to new instructions is preserved. IFR, as explained in \cite{fan2024alignformer} is shown for SQA task. [($\rightarrow$) means continued training after 38k hours.]
}
\label{tab:main_results}
\resizebox{\textwidth}{!}{
\begin{tabular}{llccccc}
\toprule
\multirow{3}{*}{Model} & \multirow{3}{*}{LLM {\tiny (Learnable) }} & \multirow{3}{1.6cm}{Train data duration {\tiny (hrs)}} & ASR {\tiny (Librispeech)} & \multicolumn{2}{
|c|}{ST {\tiny (CoVoST2)}} & SQA {\tiny (IEMOCAP)}\\
\cline{4-7}
& &  & WER $\downarrow$ & \multicolumn{2}{|c|}{BLEU $\uparrow$} & ACC $\uparrow$ \\
\cline{4-7}
& &  & t-clean / t-other & \multicolumn{1}{|c}{en-de} & \multicolumn{1}{|c|}{en-zh} & Emotion \\

\midrule
\midrule
\multicolumn{7}{c}{\textbf{Cascaded} \tiny{(CTC + LLM)} {\tiny (50k steps)}} \\

\midrule
Whisper + LLM \cite{hu2024wavllm} & LLama-7B  & - &  2.7 / 5.2 & 18.2 & - & - \\

\midrule
\multirow{2}{*}{\textbf{Ours} \quad ( Backbone )} & Phi3.1-3.8B & \multirow{2}{*}{1k} &  3.13 / 5.12 & 19.37 & 17.52 & 44.76 {\tiny (IFR = 0.99)}\\
& Qwen3-4B &  & 2.35 / 4.57 & 19.88 & 34.03 & 42.98 {\tiny (IFR = 1.00)}\\

\midrule
\midrule
\multicolumn{7}{c}{\textbf{end-to-end} {\tiny (20k steps)}} \\

\midrule
TASU \cite{peng2025tasu} & Qwen2.5-1.5B & 1k & 3.28 / 6.91 & - & 33.35 {\tiny (SFT)} & - \\
\midrule
\multirow{2}{*}{LegoSLM \cite{ma2025legoslm}} & \multirow{2}{*}{Gemma-2B {\tiny (L)}} & 56k & - / 5.2 & 21.1 {\tiny (SFT)} & -  & - \\
& & 1k & - / 7.0 & - & -  & - \\
\midrule
\multirow{2}{*}{AlignFormer \cite{fan2024alignformer}} & \multirow{2}{*}{Phi3.1-3.8B} & 38k & 3.52 / 6.47 & 15.70 & 10.97  & 31.18 {\tiny (IFR = 0.99)}\\
& & $\rightarrow$ 1k & 2.43 / 5.0 & - & -  & - \\

\midrule
\multirow{2}{*}{\textbf{Ours} \quad ( Backbone )} & Phi3.1-3.8B & \multirow{2}{*}{1k} &  2.47 / 4.71 & 16.82 & 16.49 &  37.50 {\tiny (IFR = 0.99)} \\
& Qwen3-4B & &  2.21 / 4.28 &  19.97 &  33.74 & 40.16 {\tiny (IFR = 1.00)} \\
\midrule

\multirow{1}{*}{\textbf{Ours} \quad ( Data size $\uparrow$ )} & Qwen3-4B & 2.8k & 2.48 / 4.55 & \textbf{21.13} & \textbf{36.05} & 39.60 {\tiny (IFR = 1.00)} \\
\midrule

\multirow{3}{*}{\textbf{Ours} \quad ( Model size $\uparrow$ )} & Qwen3-1.7B & \multirow{3}{*}{1k} &  2.4 / 4.43  & 14.12 & 22.87 &  29.52 {\tiny (IFR = 0.88)} \\
& Qwen3-4B & & 2.21 / 4.28 & 19.97 & 33.74 & 40.16 {\tiny (IFR = 1.00)} \\
& Qwen3-8B & &  \textbf{2.07} / \textbf{4.06} &  20.12 & 30.35 &  \textbf{41.45} {\tiny (IFR = 1.00)} \\

\bottomrule
\end{tabular}
}
\end{table*}

\subsection{Main results}

Table~\ref{tab:main_results} compares DirectSpeech2LLM against three end-to-end systems on three tasks: ASR, ST, and ER. The only seen task, ASR, during training serves as the fair comparison between end-to-end and cascaded systems. 

Despite training on only 1k hours of ASR data, DirectSpeech2LLM outperforms all comparable systems on ASR, and achieves near perfect instruction following on the two unseen tasks ST and ER (IFR = 1.00) with performance closely matching the cascaded system. A cascaded system passes discrete text to the LLM and therefore trivially preserves LLM's generalization ability. Our method approaches this upper bound zero-shot \textbf{corroborating our claim that DirectSpeech2LLM preserves the LLM's instruction following ability with speech as input, to unseen tasks}.
When evaluated using the identical \texttt{Phi3.1} backbone as AlignFormer, which enforces temporal but not geometric alignment, our method shows consistent improvement, confirming that the gains stem from the methodology rather than the LLM backbone.

DirectSpeech2LLM also shows consistent scaling behavior: increasing ASR training data from 1k to 2.8k hours further improves zero-shot translation, and scaling the LLM backbone similarly improves performance.

\renewcommand{\arraystretch}{1.0}
\begin{table*}[ht]
\centering
\caption{
Ablation study of DirectSpeech2LLM on Stage~1 and Stage~2. Assume Qwen3-4B model and LibriSpeech training data (1k) unless mentioned otherwise. 
Results compare a cascaded interface (CTC + LLM) with an end-to-end interface. 
WB and NB corresponds to With blank and No blank setup in training and inference both. Different rows show training setup and columns show inference setup. 
Hparams (rows) denote the effect of different losses, extended training steps, additional speech data (CV) or different LLM backbone. For stage 2 results, a fair comparison to Linear-CTC loss is Modified-CTC loss $- \ \mathcal{L}_{\text{commit}}$. Gray rows show the best results for each stage.
Row marked with ($\ast$) is the closest equivalent to AlignFormer minus the input ordering trick.}
\label{tab:ablations}
\resizebox{\textwidth}{!}{
\begin{tabular}{lccccc}

\toprule
\multirow{3}{*}{Hparams} & \multicolumn{3}{c|}{ASR (WER $\downarrow$)} & \multicolumn{2}{c}{ST (BLEU $\uparrow$)} \\
\cline{2-6}
& Cascaded & \multicolumn{3}{|c}{end-to-end}  & \multicolumn{1}{|c}{Cascaded} \\
\cline{2-6}
& CTC $\rightarrow$ +LLM & \multicolumn{2}{|c|}{CTC $\;/\;$ W{\tiny (ith)}B{\tiny (lank)} $\;/\;$ N{\tiny (o)}B{\tiny (lank)}} & WB $\;/\;$ NB & \multicolumn{1}{|c}{CTC + LLM} \\
\cline{2-6}
&\multicolumn{2}{c|}{Libri-other {\tiny (in-domain)}} & \multicolumn{1}{c|}{VP {\tiny (out-of-domain))}} & \multicolumn{2}{c}{CoVoST2 {\tiny (en-de)}} \\

\midrule
\midrule
\multicolumn{6}{c}{\textbf{Stage~1:} \textbf{Modified}-CTC {\tiny (20k steps)}} \\
\midrule

$\rightarrow$ \quad $\mathcal{L}_{\text{ctc}}$ & 4.88 $\rightarrow$ 4.86 & {\tiny -} / {\tiny -} / 100.00 & 16.14 / {\tiny -} / 100.00 &  {\tiny -} / 0.00 & 19.89 \\

{\transparent{0.0}$\rightarrow$} \quad $+$ \quad $\mathcal{L}_{\text{commit}}$ & 4.94 $\rightarrow$ 4.85 & {\tiny -} / {\tiny -} / 16.05 &  16.18 / {\tiny -} / 26.43 &  {\tiny -} / 10.44  & 19.92 \\

\rowcolor{gray!10} 
{\transparent{0.0}$\rightarrow$} \quad $+$ \quad 30k steps & 4.57 $\rightarrow$ 4.55 & {\tiny -} / {\tiny -} / 15.91 &  15.98 / {\tiny -} / 25.24 &  {\tiny -} / 13.57  & 19.99  \\

\midrule
$\rightarrow$ \quad $+$ CV & 4.81 $\rightarrow$ 4.80 & {\tiny -} / {\tiny -} / 16.95 &  14.06 / {\tiny -} / 24.56 &  {\tiny -} / 13.67  & 20.98  \\

\midrule
$\rightarrow$ \quad Phi3.1-3.8B & 4.62 $\rightarrow$ 10.78 & {\tiny -} / {\tiny -} / 86.48 &  15.95 / {\tiny -} / 96.23 &  {\tiny -} / 0.00  &  19.40 \\

\midrule
\midrule
\multicolumn{6}{c}{\textbf{Stage~2:} \textbf{Modified}-CTC {\tiny (50k steps) } $\;+\;$ LLM-CE {\tiny (20k steps)}  } \\
\midrule

$\rightarrow$ \quad NB & 4.63 $\rightarrow$ 4.62 & {\tiny -} / {\tiny -} / 4.63 &   15.81 / {\tiny -} / 13.64 & {\tiny -} / 19.02   & 19.92  \\

{\transparent{0.0}$\rightarrow$} \quad $-$ \quad $\mathcal{L}_{\text{commit}}$   &  4.62 $\rightarrow$ 4.61 &  {\tiny -} / {\tiny -} / 4.26 & 15.94 / {\tiny -} / 13.77 & {\tiny -} / 15.87 & 20.00 \\
\midrule

\rowcolor{gray!10} 
$\rightarrow$ \quad WB  & 4.60 $\rightarrow$ 4.57 &  {\tiny -} / 4.28 / 5.83 &  15.95 / 13.86 / 15.09 & 5.80 / 19.97 & 19.88  \\

{\transparent{0.0}$\rightarrow$} \quad $-$ \quad $\mathcal{L}_{\text{commit}}$   &  4.50 $\rightarrow$ 4.49 &  {\tiny -} / 4.10 / 5.11 & 15.92 / 13.71 / 13.61 &  5.59 /  19.69 & 19.96 \\

\midrule
$\rightarrow$ \quad WB {\tiny (decoding\_E)} & {\tiny -}  $\rightarrow$ {\tiny -} &  {\tiny -} / 7.52 / {\tiny -} & {\tiny -} / 15.55 / {\tiny -} & 19.80 / 19.97  & {\tiny -} \\
\midrule

$\rightarrow$ \quad WB $+$ \cite{fan2024alignformer} & 4.56 $\rightarrow$ 4.54 &  {\tiny -} / 4.21 / 6.26 &  15.90 / 13.73 / 15.55 &  16.04 / 19.91 & 19.90 \\

\midrule

$\rightarrow$ \quad $+$ CV & 4.88 $\rightarrow$ 4.85 & {\tiny -} / 4.55 / 7.71 & 14.18  / 12.56 / 15.01 & 5.01 / 21.13  &  21.02  \\

\midrule
$\rightarrow$ \quad Phi3.1-3.8B & 4.53 $\rightarrow$ 5.35 & {\tiny -} / 4.71 / 7.54 &  15.96 / 14.31 / 15.26 & 1.57 / 16.82 & 19.37  \\

\midrule
\midrule
\midrule
 & &  &  & & \\
\multicolumn{6}{c}{\textbf{Stage~2:} \textbf{Linear}-CTC {\tiny (50k steps)} $\;+\;$ LLM-CE {\tiny (20k steps)} }  \\
\midrule

$\rightarrow$ NB $\ast$ & 4.67 $\rightarrow$ 4.66 & {\tiny -} / {\tiny -} / 4.3 & 16.17  / {\tiny -} / 14.29 &  {\tiny -} / 1.30 & 20.00 \\
\midrule

\rowcolor{gray!10} 
$\rightarrow$ WB & 4.64 $\rightarrow$ 4.64 & {\tiny -} / 4.21 / 5.13 & 16.05 / 13.78 / 14.13 &  1.71 / 18.43 & 19.97 \\


\bottomrule
\end{tabular}
}
\end{table*}

\subsection{Ablation study}

Table~\ref{tab:ablations} ablates the key design choices of DirectSpeech2LLM and their performance on in-domain and out-of-domain seen ASR task and zero-shot unseen ST (en-de) task. We study the effects of commitment loss, scaling the training steps and data, blank token and, LLM backbone. 

\noindent\textbf{Commitment loss: }
In stage~1, the model becomes usable only once the commitment loss is used. In stage~2, removing it improves ASR performance and results in marginal degradation on unseen tasks (for results on ER; see Table \ref{tab:ext_main_results} in Section \ref{sec:appendix_hack}), suggesting that even the implicit geometric constraint within the CTC loss is sufficient to avoid prompt overfitting. Unlike Wav2Prompt which breaks on unseen tasks if the regression loss is removed as explained in Section \ref{sec:appendix_comparsionwithwav2prompt}. \textbf{This shows that geometric alignment strength plays a
smaller role than previously assumed for strong zero-shot performance on unseen tasks}. Nonetheless, we keep using the commitment loss as it consistently provides marginal gains on unseen tasks. Whether it remains necessary in a multitask setup is left for future work.

\noindent\textbf{Scaling the training steps and data: }
Increasing training steps in stage~1 from 20k to 50k and adding more (CV) data across both stages consistently improves both ASR and zero-shot translation, suggesting that more data improves robustness to out-of-domain and unseen tasks. Figure~\ref{figure:combined-ablation} (left) in Section \ref{sec:appendix_trainingsetps} further shows that ASR performance has not yet saturated in stage~2, indicating room for improvement with additional training steps.

\noindent\textbf{Role of Blank token: }
When the model is trained with no blank token (NB training) there is no prompt overfitting.

A model trained with blank token (WB training) shows poor performance on ST task not seen during training, though simply removing the blank embedding at inference fixes this. The blank token is a fixed, shared embedding (``\_'') across all training samples, so the overfitting could be \textit{positional} that is to say the model learns to associate the blank embedding's fixed position with the ASR task. This hypothesis is confirmed by the $decoding\_E$ experiment, moving the blank token to the end during inference restores instruction following. Furthermore when training with more than one task i.e.,  ASR and en-de combined\footnote{only multitask training setup in this work.}, positional overfitting of the blank embedding is substantially reduced and performance on en-zh and ER does not collapse, as shown in Table \ref{tab:multitask} in Section \ref{sec:appendix_multitas}.

Exploring the role of blank embedding to learn utterance level information such as emotion or speaker identity is left for future work.

\noindent\textbf{LLM backbone: }
The stage~1 is sensitive to the LLM backbone used, but performance gap reduces once stage~2 training kicks in. The slight under performance of \texttt{Phi3.1} compared to \texttt{Qwen3} after stage~2 training is may be because the model is weaker compared to \texttt{Qwen3} itself. 

\textbf{Lastly}, using input ordering trick, from AlignFormer, combined with our method (WB + \cite{fan2024alignformer}) reduces prompt overfitting but the performance is poor compared to our method.

\noindent\textbf{Linear CTC head: } The final two rows in Table \ref{tab:ablations} evaluate replacing our modified CTC objective with a standard linear CTC head objective. Specifically, we replace the quantization step with a learned linear layer to compute the CTC loss. In the NB training setup, this modification results in substantial degradation on the unseen ST task, consistent with the prompt overfitting observed in prior work. On the contrary in the WB training setup, the degradation is largely mitigated by removing the blank embedding during inference (NB decoding).

These observations suggest that the free-floating SFM-OES is learning task-specific cues within those embeddings, a phenomenon commonly referred to as \textit{shortcut learning} \cite{geirhos2020shortcutlearning}. In the case of WB training setup, these task-specific cues are shown to be majorly restricted to the blank token only.


\subsection{Regression loss sensitivity of Wav2Prompt vs DirectSpeech2LLM on unseen task}
\label{sec:appendix_comparsionwithwav2prompt}

\begin{table}[t]

    \centering
    \caption{The BLEU scores show the sensitivity to the MSE loss in Wav2Prompt or commitment loss in DirectSpeech2LLM. Both are regression loss for geometric alignment.}
    \label{tab:compwav2prompt}
    
    \begin{tabular}{lcc}
    \toprule
    ~  & en-es & en-de\\
    \midrule
        
    Wav2Prompt (CIF)  & 13.8 & - \\
    \quad $-$ MSE & 5.4  & - \\
    \midrule

    DirectSpeech2LLM (Modified CTC) & - & 19.97 \\
    \quad $-$ $\mathcal{L}_{\text{commit}}$ & - & 19.69 \\
    \bottomrule
    \end{tabular}
    
\end{table}

\begin{figure}[t]
\centering
\begin{tabular}{cc}
    \includegraphics[width=0.55\columnwidth]{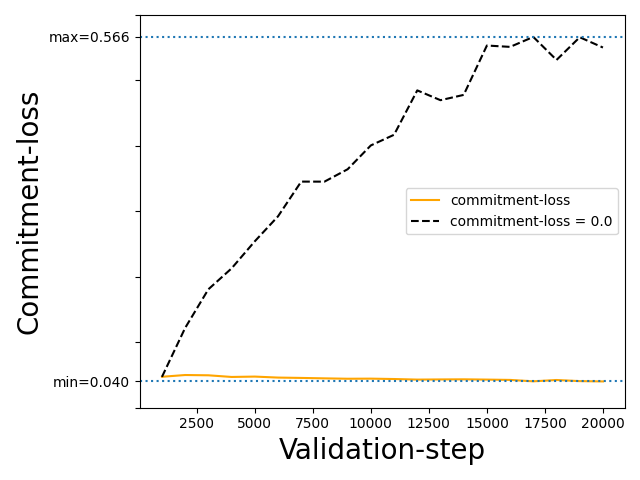} \\
\end{tabular}
\caption{Validation commitment loss of DirectSpeech2LLM with and without commitment loss.}
\label{figure:wav2promptcomparison_commitment_loss}
\end{figure}

As shown in Table~\ref{tab:compwav2prompt}, Wav2Prompt \cite{Deng2024Wav2PromptES} exhibits extreme sensitivity to its geometric alignment objective on the performance on unseen tasks, removing the MSE loss causes catastrophic degradation in zero-shot translation performance (BLEU drops from 13.8 to 5.4 for en-es language). In contrast, DirectSpeech2LLM degrades only marginally when the commitment loss is removed (19.97 $\rightarrow$ 19.69 for en-de language). 

Figure~\ref{figure:wav2promptcomparison_commitment_loss} shows this geometric divergence, despite a $10\times$ increase in embedding distances, zero-shot performance remains stable. One possible explanation is that positive-pair alignment alone does not guarantee sufficient separation between token embeddings. The LLM requires the target embeddings to remain sufficiently distinct to reliably discriminate between tokens. Thus, minimizing the distance to the positive target can be insufficient if it simultaneously reduces separation from competing embeddings. We leave a deeper investigation of this phenomenon to future work.

Since geometric alignment is already implicitly enforced in our modified CTC objective through logits computed as negative squared distances, DirectSpeech2LLM removes the need for an additional hyperparameter. Consequently, it provides a more principled mechanism that is less sensitive to hyperparameter tuning while also eliminating the train, inference mismatch introduced by CIF's length dependent weight scaling.

\section{Conclusion and Future work}
\label{sec:conclusion}

We introduce DirectSpeech2LLM, a simple end-to-end framework to align the SFM-OES to the LLM-IES of a pretrained LLM along two dimensions: temporal and geometric. This is achieved using a modified CTC objective with vector quantization. It achieves strong ASR performance and demonstrates robust instruction following on two unseen tasks, ST and ER, despite being trained solely on English ASR data. These results suggest that our method provides a simple and principled framework for integrating speech to LLM while preserving its generalization ability to unseen tasks.

\paragraph{Limitations and Future work:} Because DirectSpeech2LLM is trained on a single ASR task, its behavior in multilingual and multi-task settings remains underexplored. The role of blank and non blank embedding(s), particularly when jointly optimized with paralinguistic tasks, also remains unclear. Additionally, temporal alignment uses CTC greedy decoding, which may limit performance relative to more expressive decoding strategies, such as beam search augmented with an n-gram language model. The SFM, \texttt{Data2Vec}, is pretrained using a student-teacher setup, though we use only the student part in this work, in future we will extend the method to use teacher's pseudo labels to scale the training data in cases where we do not have the target for computing CTC loss, a semi-supervised setup. This includes cases where we have source speech and LLM targets but not source transcriptions.


\bibliographystyle{unsrtnat}
\bibliography{mybib}

@article{brown2020language,
  title={Language Models are Few-Shot Learners},
  author={Tom B. Brown and Benjamin Mann and Nick Ryder and Melanie Subbiah and Jared Kaplan and Prafulla Dhariwal and Arvind Neelakantan and Pranav Shyam and Girish Sastry and Amanda Askell and Sandhini Agarwal and Ariel Herbert-Voss and Gretchen Krueger and Thomas Henighan and Rewon Child and Aditya Ramesh and Daniel M. Ziegler and Jeff Wu and Clemens Winter and Christopher Hesse and Mark Chen and Eric Sigler and Ma-teusz Litwin and Scott Gray and Benjamin Chess and Jack Clark and Christopher Berner and Sam McCandlish and Alec Radford and Ilya Sutskever and Dario Amodei},
  journal={ArXiv},
  year={2020},
  volume={abs/2005.14165},
  url={https://api.semanticscholar.org/CorpusID:218971783}
}

@misc{touvron2023llama,
  title={LLaMA: Open and Efficient Foundation Language Models},
  author={Touvron, Hugo and Lavril, Thibaut and Sastry, Girish and others},
  howpublished={arXiv preprint arXiv:2302.13971},
  year={2023}
}

@article{openai2023gpt4,
  title={GPT-4 Technical Report},
  author={OpenAI},
  year={2023},
  url={https://api.semanticscholar.org/CorpusID:257532815}
}

@article{zhang2023speechgpt,
  title={SpeechGPT: Empowering Large Language Models with Intrinsic Cross-Modal Conversational Abilities},
  author={Zhang, Dong and Li, Shimin and Zhang, Xin and Zhan, Jun and Wang, Pengyu and Zhou, Yaqian and Qiu, Xipeng},
  journal={arXiv preprint arXiv:2305.11000},
  year={2023}
}

@article{wang2023blsp,
  title={BLSP: Bootstrapping Language-Speech Pre-training via Behavior Alignment of Continuation Writing},
  author={Wang, Chen and Liao, Minpeng and Huang, Zhongqiang and Lu, Jinliang and Wu, Junhong and Liu, Yuchen and Zong, Chengqing and Zhang, Jiajun},
  journal={arXiv preprint arXiv:2309.00916},
  year={2023}
}

@article{tang2023salmonn,
  title={SALMONN: Towards Generic Hearing Abilities for Large Language Models},
  author={Tang, Changli and Yu, Wenyi and Sun, Guangzhi and Chen, Xianzhao and Tan, Tian and Li, Wei and Lu, Lu and Ma, Zejun and Zhang, Chao},
  journal={arXiv preprint arXiv:2310.13289},
  year={2023}
}

@inproceedings{hu2024wavllm,
  title={WavLLM: Towards Robust and Adaptive Speech Large Language Model},
  author={Hu, Shujie and Zhou, Long and Liu, Shujie and Chen, Sanyuan and Meng, Lingwei and Hao, Hongkun and Pan, Jing and Liu, Xunying and Li, Jinyu and Sivasankaran, Sunit and others},
  booktitle={Findings of the Association for Computational Linguistics: EMNLP},
  pages={4552--4572},
  year={2024}
}

@misc{fan2024alignformer,
  title={AlignFormer: Modality Matching Can Achieve Better Zero-shot Instruction-Following Speech-LLM},
  author={Fan, Ruchao and Ren, Bo and Hu, Yuxuan and Zhao, Rui and Liu, Shujie and Li, Jinyu},
  howpublished={arXiv preprint arXiv:2412.01145},
  year={2024}
}

@article{peng2025tasu,
  title={TASU: Text-Only Alignment for Speech Understanding},
  author={Peng, Jing and Yang, Yi and Li, Xu and Xi, Yu and Tang, Quanwei and Fang, Yangui and Li, Junjie and Yu, Kai},
  journal={arXiv preprint arXiv:2511.03310},
  year={2025}
}

@article{ma2025legoslm,
  title={LegoSLM: Connecting LLM with Speech Encoder Using CTC Posteriors},
  author={Ma, Rao and Chen, Tongzhou and Audhkhasi, Kartik and Ramabhadran, Bhuvana},
  journal={arXiv preprint arXiv:2505.11352},
  year={2025}
}

@article{mohapatra2026speechmapper,
  title={SpeechMapper: Speech-to-text Embedding Projector for LLMs},
  author={Mohapatra, Biswesh and Boito, Marcely Zanon and Calapodescu, Ioan},
  journal={arXiv preprint arXiv:2601.20417},
  year={2026}
}

@inproceedings{graves2006ctc,
  title={Connectionist Temporal Classification: Labelling Unsegmented Sequence Data with Recurrent Neural Networks},
  author={Graves, Alex and Fern{\'a}ndez, Santiago and Gomez, Faustino and Schmidhuber, J{\"u}rgen},
  booktitle={Proceedings of the 23rd International Conference on Machine Learning},
  pages={369--376},
  year={2006}
}

@article{van2017neuralvqvae,
  title={Neural Discrete Representation Learning},
  author={van den Oord, Aaron and Vinyals, Oriol and Kavukcuoglu, Koray},
  journal={Advances in Neural Information Processing Systems},
  volume={30},
  year={2017}
}

@inproceedings{wang-etal-2021-voxpopuli,
    title = "{V}ox{P}opuli: A Large-Scale Multilingual Speech Corpus for Representation Learning, Semi-Supervised Learning and Interpretation",
    author = "Wang, Changhan  and
      Riviere, Morgane  and
      Lee, Ann  and
      Wu, Anne  and
      Talnikar, Chaitanya  and
      Haziza, Daniel  and
      Williamson, Mary  and
      Pino, Juan  and
      Dupoux, Emmanuel",
    booktitle = "Proceedings of the 59th Annual Meeting of the Association for Computational Linguistics and the 11th International Joint Conference on Natural Language Processing (Volume 1: Long Papers)",
    month = aug,
    year = "2021",
    address = "Online",
    publisher = "Association for Computational Linguistics",
    url = "https://aclanthology.org/2021.acl-long.80",
    pages = "993--1003",
}

@inproceedings{panayotov2015librispeech,
  title={Librispeech: an ASR corpus based on public domain audio books},
  author={Panayotov, Vassil and Chen, Guoguo and Povey, Daniel and Khudanpur, Sanjeev},
  booktitle={Acoustics, Speech and Signal Processing (ICASSP), 2015 IEEE International Conference on},
  pages={5206--5210},
  year={2015},
  organization={IEEE}
}

@misc{wang2020covost2,
    title={CoVoST 2: A Massively Multilingual Speech-to-Text Translation Corpus},
    author={Changhan Wang and Anne Wu and Juan Pino},
    year={2020},
    eprint={2007.10310},
    archivePrefix={arXiv},
    primaryClass={cs.CL}
}

@article{busso2008iemocap,
  title={IEMOCAP: Interactive emotional dyadic motion capture database},
  author={Busso, Carlos and Bulut, Murtaza and Lee, Chi-Chun and Kazemzadeh, Abe and Mower, Emily and Kim, Samuel and Chang, Jeannette N and Lee, Sungbok and Narayanan, Shrikanth S},
  journal={Language resources and evaluation},
  volume={42},
  number={4},
  pages={335--359},
  year={2008},
  publisher={Springer}
}

@inproceedings{ardila2020common,
  title={Common voice: A massively-multilingual speech corpus},
  author={Ardila, Rosana and Branson, Megan and Davis, Kelly and Kohler, Michael and Meyer, Josh and Henretty, Michael and Morais, Reuben and Saunders, Lindsay and Tyers, Francis and Weber, Gregor},
  booktitle={Proceedings of the twelfth language resources and evaluation conference},
  pages={4218--4222},
  year={2020}
}

@inproceedings{baevski2022data2vec,
  title={Data2vec: A general framework for self-supervised learning in speech, vision and language},
  author={Baevski, Alexei and Hsu, Wei-Ning and Xu, Qiantong and Babu, Arun and Gu, Jiatao and Auli, Michael},
  booktitle={International conference on machine learning},
  pages={1298--1312},
  year={2022},
  organization={PMLR}
}

@inproceedings{pagnoni2025blt,
  title={Byte latent transformer: Patches scale better than tokens},
  author={Pagnoni, Artidoro and Pasunuru, Ramakanth and Rodriguez, Pedro and Nguyen, John and Muller, Benjamin and Li, Margaret and Zhou, Chunting and Yu, Lili and Weston, Jason E and Zettlemoyer, Luke and others},
  booktitle={Proceedings of the 63rd Annual Meeting of the Association for Computational Linguistics (Volume 1: Long Papers)},
  pages={9238--9258},
  year={2025}
}

@inproceedings{wu2023decoderctcdownsample1,
  title={On decoder-only architecture for speech-to-text and large language model integration},
  author={Wu, Jian and Gaur, Yashesh and Chen, Zhuo and Zhou, Long and Zhu, Yimeng and Wang, Tianrui and Li, Jinyu and Liu, Shujie and Ren, Bo and Liu, Linquan and others},
  booktitle={2023 IEEE automatic speech recognition and understanding workshop (ASRU)},
  pages={1--8},
  year={2023},
  organization={IEEE}
}

@article{pan2023cosmicspeechintsrtcutiontunign,
  title={Cosmic: Data efficient instruction-tuning for speech in-context learning},
  author={Pan, Jing and Wu, Jian and Gaur, Yashesh and Sivasankaran, Sunit and Chen, Zhuo and Liu, Shujie and Li, Jinyu},
  journal={arXiv preprint arXiv:2311.02248},
  year={2023}
}

@article{hsu2021hubert,
  title={Hubert: Self-supervised speech representation learning by masked prediction of hidden units},
  author={Hsu, Wei-Ning and Bolte, Benjamin and Tsai, Yao-Hung Hubert and Lakhotia, Kushal and Salakhutdinov, Ruslan and Mohamed, Abdelrahman},
  journal={IEEE/ACM transactions on audio, speech, and language processing},
  volume={29},
  pages={3451--3460},
  year={2021},
  publisher={IEEE}
}

@article{zeghidour2021soundstream,
  title={Soundstream: An end-to-end neural audio codec},
  author={Zeghidour, Neil and Luebs, Alejandro and Omran, Ahmed and Skoglund, Jan and Tagliasacchi, Marco},
  journal={IEEE/ACM Transactions on Audio, Speech, and Language Processing},
  volume={30},
  pages={495--507},
  year={2021},
  publisher={IEEE}
}

@inproceedings{radford2023robustwhisper,
  title={Robust speech recognition via large-scale weak supervision},
  author={Radford, Alec and Kim, Jong Wook and Xu, Tao and Brockman, Greg and McLeavey, Christine and Sutskever, Ilya},
  booktitle={International conference on machine learning},
  pages={28492--28518},
  year={2023},
  organization={PMLR}
}

@inproceedings{koh2021wilds,
  title={Wilds: A benchmark of in-the-wild distribution shifts},
  author={Koh, Pang Wei and Sagawa, Shiori and Marklund, Henrik and Xie, Sang Michael and Zhang, Marvin and Balsubramani, Akshay and Hu, Weihua and Yasunaga, Michihiro and Phillips, Richard Lanas and Gao, Irena and others},
  booktitle={International conference on machine learning},
  pages={5637--5664},
  year={2021},
  organization={PMLR}
}

@article{liu2023languagefixedcodebook1,
  title={Language quantized autoencoders: Towards unsupervised text-image alignment},
  author={Liu, Hao and Yan, Wilson and Abbeel, Pieter},
  journal={Advances in neural information processing systems},
  volume={36},
  pages={4382--4395},
  year={2023}
}

@article{mentzer2023finitefsqfixedcodebook,
  title={Finite scalar quantization: Vq-vae made simple},
  author={Mentzer, Fabian and Minnen, David and Agustsson, Eirikur and Tschannen, Michael},
  journal={arXiv preprint arXiv:2309.15505},
  year={2023}
}

@article{xue2022byt5,
  title={ByT5: Towards a token-free future with pre-trained byte-to-byte models},
  author={Xue, Linting and Barua, Aditya and Constant, Noah and Al-Rfou, Rami and Narang, Sharan and Kale, Mihir and Roberts, Adam and Raffel, Colin},
  journal={Transactions of the Association for Computational Linguistics},
  volume={10},
  pages={291--306},
  year={2022},
  publisher={MIT Press One Broadway, 12th Floor, Cambridge, Massachusetts 02142, USA~…}
}

@inproceedings{yu2023megabyte,
  title     = {Megabyte: Predicting Million-Byte Sequences with Multiscale Transformers},
  author    = {Yu, Lili and Simig, Dániel and Flaherty, Colin and Aghajanyan, Armen and Zettlemoyer, Luke and Lewis, Mike},
  booktitle = {Advances in Neural Information Processing Systems},
  volume    = {36},
  pages     = {78808--78823},
  year      = {2023}
}

@misc{chen2025audiollmsreallylisten,
      title={Do Audio LLMs Really LISTEN, or Just Transcribe? Measuring Lexical vs. Acoustic Emotion Cues Reliance}, 
      author={Jingyi Chen and Zhimeng Guo and Jiyun Chun and Pichao Wang and Andrew Perrault and Micha Elsner},
      year={2025},
      eprint={2510.10444},
      archivePrefix={arXiv},
      primaryClass={cs.CL},
      url={https://arxiv.org/abs/2510.10444}, 
}

@misc{gao2022wavpromptfewshotspokenlanguage,
      title={WAVPROMPT: Towards Few-Shot Spoken Language Understanding with Frozen Language Models}, 
      author={Heting Gao and Junrui Ni and Kaizhi Qian and Yang Zhang and Shiyu Chang and Mark Hasegawa-Johnson},
      year={2022},
      eprint={2203.15863},
      archivePrefix={arXiv},
      primaryClass={eess.AS},
      url={https://arxiv.org/abs/2203.15863}, 
}

@article{Peng_2026_survey_speech_llm,
   title={A Survey on Speech Large Language Models for Understanding},
   volume={20},
   ISSN={1941-0484},
   url={http://dx.doi.org/10.1109/JSTSP.2025.3640535},
   DOI={10.1109/jstsp.2025.3640535},
   number={1},
   journal={IEEE Journal of Selected Topics in Signal Processing},
   publisher={Institute of Electrical and Electronics Engineers (IEEE)},
   author={Peng, Jing and Wang, Yucheng and Li, Bohan and Guo, Yiwei and Wang, Hankun and Fang, YanGui and Xi, Yu and Li, Haoyu and Li, Xu and Zhang, Ke and Wang, Shuai and Yu, Kai},
   year={2026},
   month=jan, pages={2–31} }

@misc{chu2023qwenaudioadvancinguniversalaudioqwen1,
      title={Qwen-Audio: Advancing Universal Audio Understanding via Unified Large-Scale Audio-Language Models}, 
      author={Yunfei Chu and Jin Xu and Xiaohuan Zhou and Qian Yang and Shiliang Zhang and Zhijie Yan and Chang Zhou and Jingren Zhou},
      year={2023},
      eprint={2311.07919},
      archivePrefix={arXiv},
      primaryClass={eess.AS},
      url={https://arxiv.org/abs/2311.07919}, 
}

@misc{cuervo2026closinggaptextspeech,
      title={Closing the Gap Between Text and Speech Understanding in LLMs}, 
      author={Santiago Cuervo and Skyler Seto and Maureen de Seyssel and Richard He Bai and Zijin Gu and Tatiana Likhomanenko and Navdeep Jaitly and Zakaria Aldeneh},
      year={2026},
      eprint={2510.13632},
      archivePrefix={arXiv},
      primaryClass={cs.CL},
      url={https://arxiv.org/abs/2510.13632}, 
}

@misc{zeng2024scalingspeechtextpretrainingsyntheticmoredata,
      title={Scaling Speech-Text Pre-training with Synthetic Interleaved Data}, 
      author={Aohan Zeng and Zhengxiao Du and Mingdao Liu and Lei Zhang and Shengmin Jiang and Yuxiao Dong and Jie Tang},
      year={2024},
      eprint={2411.17607},
      archivePrefix={arXiv},
      primaryClass={cs.CL},
      url={https://arxiv.org/abs/2411.17607}, 
}

@misc{li2021prefixtuningoptimizingcontinuousprompts,
      title={Prefix-Tuning: Optimizing Continuous Prompts for Generation}, 
      author={Xiang Lisa Li and Percy Liang},
      year={2021},
      eprint={2101.00190},
      archivePrefix={arXiv},
      primaryClass={cs.CL},
      url={https://arxiv.org/abs/2101.00190}, 
}

@inproceedings{oh2026seamrobustness2,
  title={SEAM: Bridging the Temporal-Semantic Granularity Gap for LLM-based Speech Recognition},
  author={Oh, Junseok and Kim, Ji-Hwan},
  booktitle={Findings of the Association for Computational Linguistics: EACL 2026},
  pages={2135--2144},
  year={2026}
}

@inproceedings{kumar2025performancerobustness1,
  title={Performance evaluation of slam-asr: The good, the bad, the ugly, and the way forward},
  author={Kumar, Shashi and Thorbecke, Iuliia and Burdisso, Sergio and Villatoro-Tello, Esa{\'u} and KE, Manjunath and Hacio{\u{g}}lu, Kadri and Rangappa, Pradeep and Motlicek, Petr and Ganapathiraju, Aravind and Stolcke, Andreas},
  booktitle={2025 IEEE International Conference on Acoustics, Speech, and Signal Processing Workshops (ICASSPW)},
  pages={1--5},
  year={2025},
  organization={IEEE}
}

@misc{ma2024embarrassinglysimpleapproachllmslamasr,
      title={An Embarrassingly Simple Approach for LLM with Strong ASR Capacity}, 
      author={Ziyang Ma and Guanrou Yang and Yifan Yang and Zhifu Gao and Jiaming Wang and Zhihao Du and Fan Yu and Qian Chen and Siqi Zheng and Shiliang Zhang and Xie Chen},
      year={2024},
      eprint={2402.08846},
      archivePrefix={arXiv},
      primaryClass={cs.CL},
      url={https://arxiv.org/abs/2402.08846}, 
}

@article{yadav2024jooci,
  title={JOOCI: a Framework for Learning Comprehensive Speech Representations},
  author={Yadav, Hemant and Shah, Rajiv Ratn and Sitaram, Sunayana},
  journal={arXiv preprint arXiv:2410.11086},
  year={2024}
}

@article{liu2023dinosr,
  title={Dinosr: Self-distillation and online clustering for self-supervised speech representation learning},
  author={Liu, Alexander H and Chang, Heng-Jui and Auli, Michael and Hsu, Wei-Ning and Glass, Jim},
  journal={Advances in Neural Information Processing Systems},
  volume={36},
  pages={58346--58362},
  year={2023}
}

@inproceedings{zhang2025soundwave,
  title={Soundwave: Less is more for speech-text alignment in llms},
  author={Zhang, Yuhao and Liu, Zhiheng and Bu, Fan and Zhang, Ruiyu and Wang, Benyou and Li, Haizhou},
  booktitle={Proceedings of the 63rd Annual Meeting of the Association for Computational Linguistics (Volume 1: Long Papers)},
  pages={18718--18738},
  year={2025}
}

@inproceedings{tsiamas2024pushingspeechtransaltionctclaignnmentoptimaltransport2,
  title={Pushing the limits of zero-shot end-to-end speech translation},
  author={Tsiamas, Ioannis and G{\'a}llego, Gerard I and Fonollosa, Jos{\'e} AR and Costa-juss{\`a}, Marta R},
  booktitle={Findings of the Association for Computational Linguistics: ACL 2024},
  pages={14245--14267},
  year={2024}
}

@inproceedings{le2023prespeechtransaltionctclaignnmentoptimaltransport2,
  title={Pre-training for speech translation: Ctc meets optimal transport},
  author={Le, Phuong-Hang and Gong, Hongyu and Wang, Changhan and Pino, Juan and Lecouteux, Benjamin and Schwab, Didier},
  booktitle={International Conference on Machine Learning},
  pages={18667--18685},
  year={2023},
  organization={PMLR}
}

@article{Caron2021EmergingPIDINO,
  title={Emerging Properties in Self-Supervised Vision Transformers},
  author={Mathilde Caron and Hugo Touvron and Ishan Misra and Herv'e J'egou and Julien Mairal and Piotr Bojanowski and Armand Joulin},
  journal={2021 IEEE/CVF International Conference on Computer Vision (ICCV)},
  year={2021},
  pages={9630-9640},
  url={https://api.semanticscholar.org/CorpusID:233444273}
}

@article{Park2019SpecAugmentAS,
  title={SpecAugment: A Simple Data Augmentation Method for Automatic Speech Recognition},
  author={Daniel S. Park and William Chan and Yu Zhang and Chung-Cheng Chiu and Barret Zoph and Ekin Dogus Cubuk and Quoc V. Le},
  journal={ArXiv},
  year={2019},
  volume={abs/1904.08779},
  url={https://api.semanticscholar.org/CorpusID:121321299}
}

@article{Deng2024Wav2PromptES,
  title={Wav2Prompt: End-to-End Speech Prompt Generation and Tuning For LLM in Zero and Few-shot Learning},
  author={Keqi Deng and Guangzhi Sun and Phil Woodland},
  journal={ArXiv},
  year={2024},
  volume={abs/2406.00522},
  url={https://api.semanticscholar.org/CorpusID:270220534}
}

@article{geirhos2020shortcutlearning,
  title={Shortcut learning in deep neural networks},
  author={Geirhos, Robert and Jacobsen, J{\"o}rn-Henrik and Michaelis, Claudio and Zemel, Richard and Brendel, Wieland and Bethge, Matthias and Wichmann, Felix A},
  journal={Nature Machine Intelligence},
  volume={2},
  number={11},
  pages={665--673},
  year={2020},
  publisher={Nature Publishing Group UK London}
}

\clearpage
\appendix
\section{Appendix}
\label{sec:appendix}

\subsection{Preliminaries}
\label{sec:appendix_background}

\subsubsection{Large Language Model (LLM) Input}
LLMs process language by mapping raw text into sequences of discrete tokens using a fixed set of rules (temporal alignment). Each token is then mapped to a high-dimensional embedding (geometric alignment). Together, these define two dimensions along which the input can deviate from the seen input distribution: the token sequence and the embedding space. While LLMs tolerate small perturbations, changes to either can significantly degrade performance \cite{mohapatra2026speechmapper, koh2021wilds}, resulting in a loss of its generalization to follow instructions.

Therefore, constructing an effective Speech-LLM requires satisfying two alignment criteria: temporal and geometric. \textbf{Our hypothesis} is that minimizing distribution along these two dimensions \textbf{may} improve cross-modal alignment resulting in increased generalization to unseen tasks.

\subsubsection{Connectionist Temporal Classification (CTC)} 
CTC \cite{graves2006ctc} is a sequence learning objective designed for tasks where the exact alignment between input frames and output tokens is unknown, such as ASR. Given an input sequence of acoustic representations $\mathbf{X}$ and a target token sequence $\mathbf{Y}$, CTC models the conditional probability by marginalizing over all possible frame level alignments that collapse to $\mathbf{Y}$. To enable flexible alignment, CTC introduces a special \textit{blank} symbol that allows the model to emit no valid token at a given timestep. During inference, repeated tokens and blanks are removed through a collapsing operation to produce the final transcription.

\subsubsection{Vector Quantization (VQ)} 
VQ \cite{van2017neuralvqvae} is a process of committing continuous representations to a discrete codebook, which in this work is identified with the LLM input embedding space (IES). Given an SFM output $\mathbf{z}$, the quantizer selects the nearest codebook vector $\mathbf{q}_k \in \mathcal{E}={\mathbf{q}_1,\dots,\mathbf{q}_K}$, encouraging the representation to align with its quantized counterpart before passing it downstream. The use of vector quantization in speech \cite{hsu2021hubert, zeghidour2021soundstream} or in vision \cite{Caron2021EmergingPIDINO} representation learning  is well established. However, existing methods primarily treat the codebook as an auxiliary component for generating pseudo-labels, or learn it jointly during training \cite{liu2023dinosr}. 

Unlike standard VQ, which learns a dynamic codebook, we instead fix the codebook to be LLM-IES, following approaches such as \cite{liu2023languagefixedcodebook1, mentzer2023finitefsqfixedcodebook}. We reinterpret quantization as a classification problem, where mean squared error scores are used as logits, reducing the quantizer to a direct prediction layer over codebook indices, providing a stable interface for end-to-end integration between speech and the language model.

\subsection{Training Procedure}
\label{sec:appendix_tp}
\noindent\textbf{Stage 1: }
The SFM is trained using the modified CTC objective and the commitment loss as described in Section~\ref{sec:methods}. 
During this stage, the last six layers of the \texttt{Data2Vec-Large}  are finetuned. Training runs for 50k steps and the peak learning rate (LR) is $1 \times 10^{-4}$ . The LR is linearly warm-up for the first $10\%$ of the steps and then linearly decayed for the remaining steps. 

\noindent\textbf{Stage 2: }
We add cross-entropy loss from LLM to stage 1 losses, while keeping the LLM frozen. At this stage, only the last two layers of the \texttt{Data2Vec-Large}  encoder are finetuned. Training runs for 20k steps and the peak LR is $5 \times 10^{-5}$ . The LR is linearly warm-up for the first $10\%$ of the steps and kept constant for $80\%$ of the steps and then linearly decay for the remaining steps. 

Both stages use the AdamW optimizer with gradient clipping set to $1$ using 4 B200 GPUs with 180 GB memory. Each GPU processes at max 200 seconds of audio per step and the total batch has at max 800 seconds. All experiments are conducted using \textit{TF32} precision. Stage 1 (50k steps) completes in approximately 2.5 hours and Stage 2 (20k steps) in approximately 2 hours for the Qwen3-4B configuration.


\subsection{Multitask Training Setup}
\label{sec:appendix_multitas}

Table \ref{tab:multitask} reports results under a multitask training configuration in which the model is jointly trained on LibriSpeech ASR data and English-to-German speech translation (en-de) data from CoVoST2. Commitment loss is excluded in this setup to isolate the effect of multitask supervision.

The primary finding is that positional overfitting to the blank token embedding, a limitation of the single task WB setup is substantially reduced when more than one task is present during training. The model learns to treat the blank token as a general purpose utterance level embedding rather than a fixed cue for the ASR task.
Notably, the WB setup in the multitask configuration outperforms the cascaded system on all seen tasks. This indicates that end-to-end training with a contextual blank embedding is advantageous over the cascaded pipeline.

\renewcommand{\arraystretch}{1.0}
\begin{table}[h]
\centering
\caption{Multitask training setup with Librispeech and en-de training data without commitment loss. }
\label{tab:multitask}
\begin{tabular}{lcccc}
\toprule
Tasks & WB & WB\tiny{decoding\_E} & NB & CTC + LLM\\
\midrule

Libri-clean & 2.4 & 2.95 & 2.8  & 2.39 \\
Libri-other & 4.39 & 5.73 & 5.51  & 4.61 \\
en-de & 21.45 & 21.61 & 21.68  & 20.29 \\
en-zh & 34.20 & 33.91  & 35.23  & 34.98 \\
Emotion & 40.89 & 39.35 & 41.53  & 43.55\\

\bottomrule
\end{tabular}
\end{table}

\subsection{Longform Speech Inference}
\label{sec:appendix_longform}

We evaluate DirectSpeech2LLM on longform speech by concatenating LibriSpeech test-clean utterances of duration ranging from 1 to 20 minutes. The model is trained on utterances of upto 20 seconds; evaluation on longer sequences thus constitutes out-of-distribution inference with respect to sequence length.

The results in Table \ref{tab:appendix_longform} reveal a clear degradation pattern as duration increases. At 1 minute, both WB and NB setups match the CTC baseline. By 10 minutes, WB degrades substantially (WER 14.26), while NB remains comparatively robust (WER 4.22). At 20 minutes, all setups degrade significantly.

This behavior reflects a well known limitation of transformer self-attention, attention patterns do not extrapolate reliably to sequence lengths beyond the training distribution. The blank embedding in the WB setup compounds this issue by encoding a fixed positional cue that becomes increasingly misaligned to longer durations. Restricting the SFM to a fixed sliding-window attention mechanism (e.g., 20-60 second windows) is a natural remedy that we leave this to future work.

\renewcommand{\arraystretch}{1.0}
\begin{table}[ht]
\centering
\caption{Longform speech inference results. }
\label{tab:appendix_longform}
\begin{tabular}{lccc}
\toprule
Duration & WB & NB & CTC \\
\midrule

1 min & .62 & 0.62 & 0.62  \\
5 min & 2.97 & 2.6 & 3.71 \\
10 min & 14.26 & 4.22 & 5.08 \\
20 min & 38.48 & 33.12 & 12.37\\

\bottomrule
\end{tabular}
\end{table}

\subsection{Training dynamics}
\label{sec:appendix_trainingsetps}

Figure~\ref{figure:combined-ablation} presents two views of training dynamics across the Qwen3-4B and Phi-3.1-3.8B LLM backbones.

\subsubsection{WER Convergence (Figure~\ref{figure:combined-ablation}, Left)}
\begin{figure}[t]
\centering
\begin{tabular}{cc}
    \includegraphics[width=0.45\columnwidth]{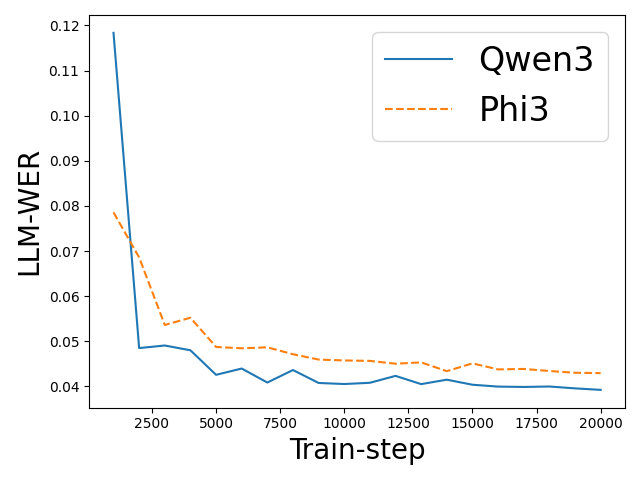} &
    \includegraphics[width=0.45\columnwidth]{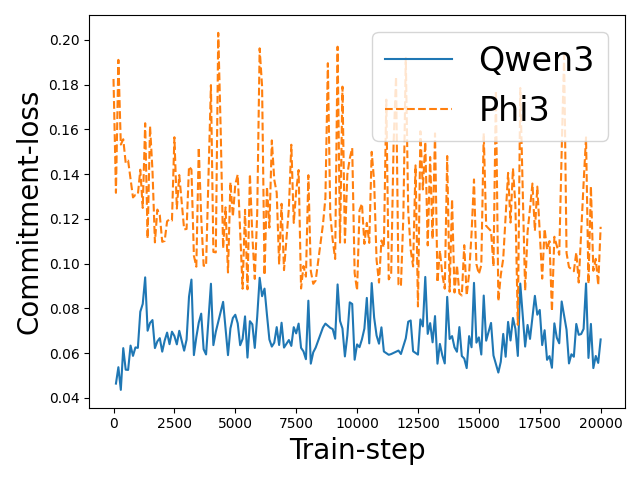} \\
\end{tabular}
\caption{Left: Stage~2 LLM WER vs LLM backbone shows training longer would improve performance further. Right: Commitment loss vs LLM backbone.}
\label{figure:combined-ablation}
\end{figure}

The left panel shows Stage 2 LLM WER on LibriSpeech test-other as a function of training step. Both backbones exhibit sustained improvement throughout the 20,000 step training budget, with no sign of convergence plateauing. These trends indicate that performance gains from extended training are likely available for both backbones, and that our reported results represent a lower bound on what is achievable with the current framework.
\subsubsection{Commitment Loss (Figure~\ref{figure:combined-ablation}, Right)}
The right panel shows the commitment loss trajectory during Stage 2 training. Phi-3.1-3.8B exhibits substantially higher and more erratic commitment loss values compared to Qwen3-4B, which converges to a consistently lower loss with less variance. This indicates that the SFM outputs align more easily with the Qwen3-4B embeddings geometry. The underlying cause of this difference whether it stems from the size, training data, or architectural characteristics of each LLM remains an open question that we leave for future investigation.

Taken together, these training dynamics reinforce two conclusions from the main paper: (i) DirectSpeech2LLM's performance has not saturated and stands to benefit from longer training, and (ii) LLM backbone quality is a meaningful factor in the geometric alignment between the SFM output space and the LLM input embedding space.

\clearpage
\subsection{Token Boundary Preservation}
\label{sec:appendix_hack}

\begin{figure}[t]
\centering
\includegraphics[width=\textwidth, trim=10cm 8cm 5.8cm 7cm, clip]{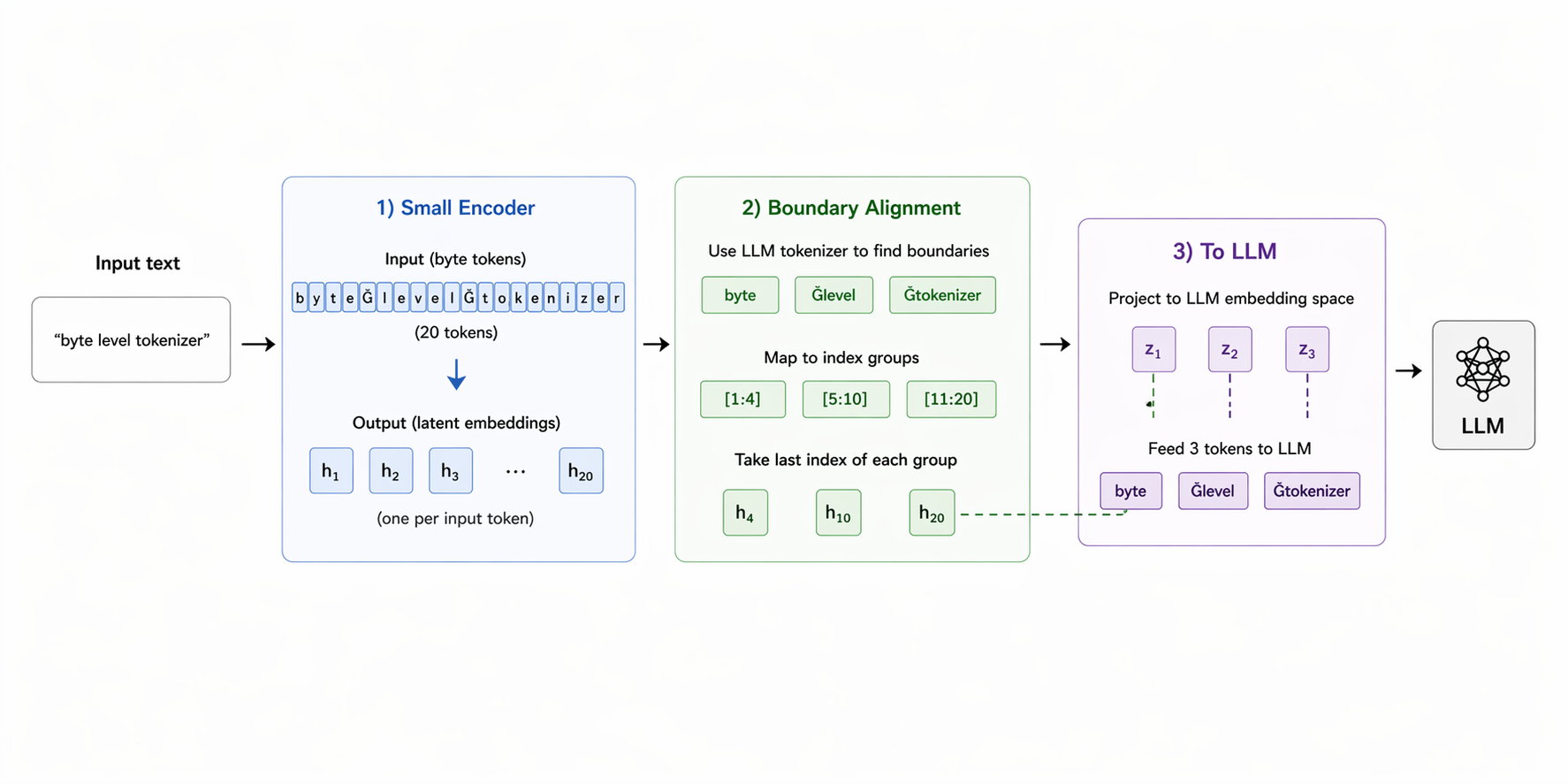}
\caption{TBP method overview. The authors of Byte Latent Transformer ~\cite{pagnoni2025blt} use dynamic boundaries at step 2, while TBP uses static tokenizer dependent boundaries.}
\label{figure:hack}
\end{figure}

We propose Token Boundary Preservation (TBP), implemented via a Bytes-to-Token Causal encoder (BTC) that aggregates byte-level representations according to tokenizer-specific token spans.

Consider the input text \texttt{"byte level tokenizer"}. Qwen3 LLM tokenizer splits this string into 3 tokens:
\[
[\text{byte},\ \text{Ġlevel},\ \text{Ġtokenizer}].
\]

In contrast, the Speech-LLM tokenizer of DirectSpeech2LLM method operating on a restricted vocabulary of $1$K splits this string into 9 tokens:
\[
[\text{by},\ \text{te},\ \text{Ġle},\ \text{vel},\ \text{Ġto},\ \text{k},\ \text{en},\ \text{iz},\ \text{er}].
\]

Similarly, a pure byte-level tokenizer splits this string into 20 tokens:
\[
[b,\ y,\ t,\ e,\ \text{Ġ},\ l,\ e,\ v,\ e,\ l,\ \text{Ġ},\ t,\ o,\ k,\ e,\ n,\ i,\ z,\ e,\ r].
\]

Our objective is to preserve the tokenization spans seen by the LLM during pre-training (i.e., the original 3 tokens) rather than feeding different sequences of 9 or 20 tokens. TBP uses the LLM's tokenizer to define deterministic \emph{token spans} over byte-level representations, different from BLT's stochastic spans which introduces additional complexity. Formally, let the BTC encoder process the byte-level sequence:
\[
[b,\ y,\ t,\ e,\ \text{Ġ},\ l,\ e,\ v,\ e,\ l,\ \text{Ġ},\ t,\ o,\ k,\ e,\ n,\ i,\ z,\ e,\ r]
\]
to produce latent embeddings $\{h_1, \dots, h_{20}\}$. Applying the LLM tokenizer to the original text yields token spans corresponding to:
\[
[\text{byte},\ \text{Ġlevel},\ \text{Ġtokenizer}],
\]
which map to byte-index ranges $[1\!:\!4]$, $[5\!:\!10]$, and $[11\!:\!20]$. Using these spans, the BTC encoder aggregates representations by selecting the final latent vector from each span:
\[
\{h_4,\ h_{10},\ h_{20}\}.
\]

\begin{figure}[ht]
\centering
\includegraphics[width=\textwidth]{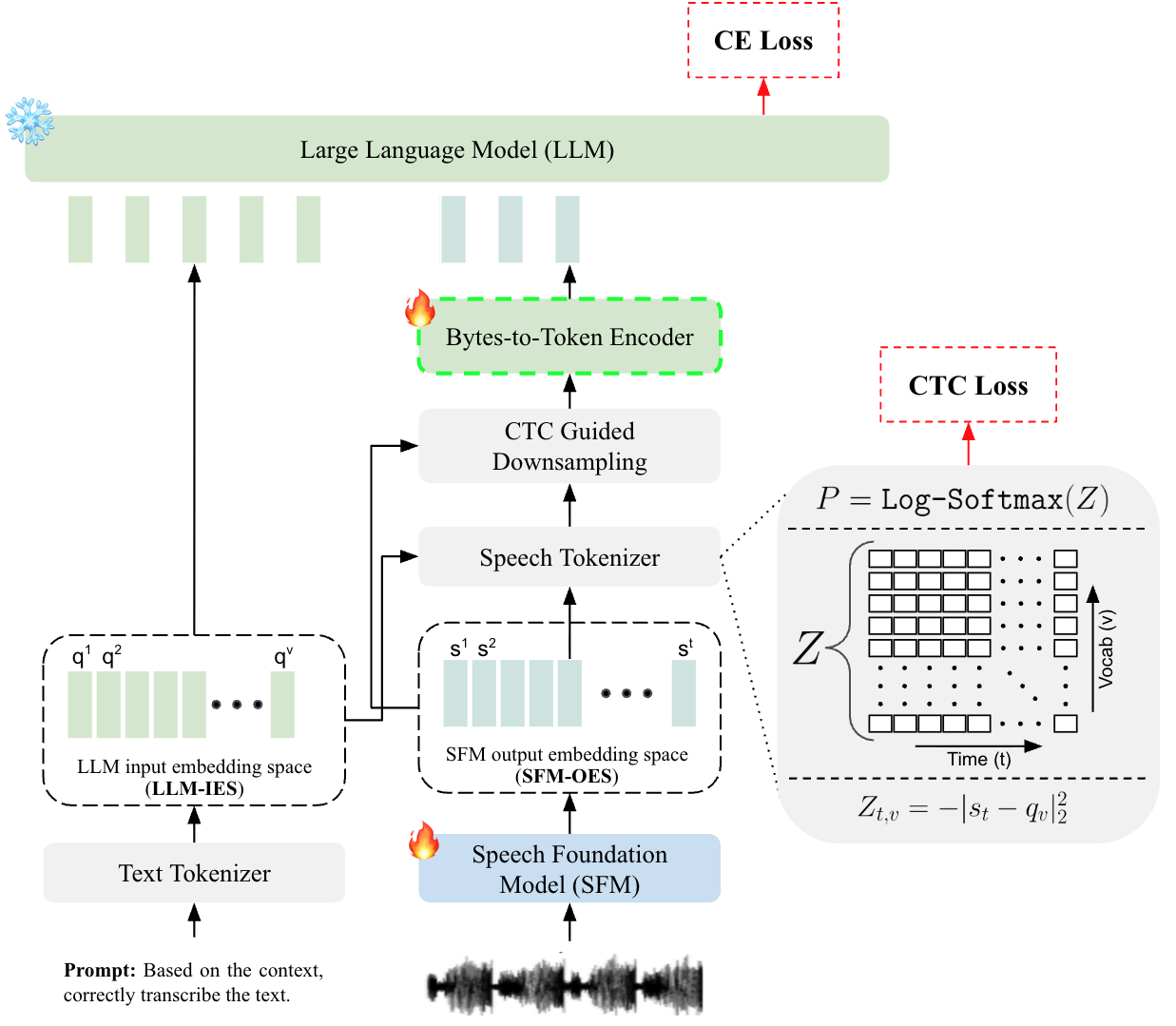}
\caption{Updated DirectSpeech2LLM\_TBP method.}
\label{figure:BA_method}
\end{figure}

These representations are then projected into the LLM input embedding space and passed as input to the LLM. To ensure compatibility with the LLM input embedding space (IES), BTC encoder logits are calculated similar to the Stage 1 of DirectSpeech2LLM but instead of CTC we use cross-entropy loss. An overview of the TBP is illustrated in Figure~\ref{figure:hack} and the updated DirectSpeech2LLM\_TBP method is illustrated in Figure~\ref{figure:BA_method}. The results are shown in Table \ref{tab:ext_main_results} and \ref{tab:ext_ablations}.

Because BTC embeddings are conditioned on raw bytes, they might be more robust to spelling errors and better at tasks which require character-level granularity, such as counting the "r"s in "strawberry" or solving math. Furthermore, we plan to explore using multiple TBP rules with prefix tuning, for a single BTC. In multilingual settings, this would allow us to dynamically choose the TBP rule that produces the fewest tokens for a given language.

\renewcommand{\arraystretch}{1.1}
\begin{table}[t]
\centering
\caption{
Extension to Table \ref{tab:main_results}. Numbers with \underline{underline} shows the best result for each test set. The improvements except on the ASR task are marginal.  Overall, the results suggest that input tokenization has no substantial effect on the LLM’s task-solving ability. Values in parentheses report ASR results computed using the Whisper EnglishTextNormalizer compared to mms normalization.}
\label{tab:ext_main_results}
\begin{tabular}{llccccc}
\toprule
- & - & - & t-clean / t-other & en-de & en-zh & Emotion \\


\midrule
\multicolumn{7}{c}{\textbf{end-to-end}} \\

\hline
Ours  & \multirow{3}{*}{Qwen-3-4B} & \multirow{3}{*}{1k} & 2.21 (2.05) / 4.28 (4.13) & \underline{19.97} & \underline{33.74} &  40.16 {\tiny (IFR = 1.00)} \\
Ours $-$ $\mathcal{L}_{\text{commit}}$ & & &  2.13 (1.96) / 4.09 (3.93) & 19.69 & 32.86 & 39.52 {\tiny (IFR = 1.00)} \\

Ours $+$ TBP &  & & \underline{1.96} (1.81) / \underline{4.06} (3.90) & 19.79 & 32.74 & \underline{40.97} {\tiny (IFR = 1.00)}\\

\bottomrule
\end{tabular}
\end{table}

\renewcommand{\arraystretch}{1.1}
\begin{table}[t]
\centering
\caption{Extension to Table \ref{tab:ablations}. Numbers with \underline{underline} shows the best result for each test set.}
\label{tab:ext_ablations}
\resizebox{\textwidth}{!}{
\begin{tabular}{lccccc}
\toprule
\multirow{3}{*}{Hparams} & \multicolumn{3}{c|}{ASR (WER $\downarrow$)} & \multicolumn{2}{c}{ST (BLEU $\uparrow$)} \\
\cline{2-6}
& Cascaded & \multicolumn{3}{|c}{end-to-end}  & \multicolumn{1}{|c}{Cascaded} \\
\cline{2-6}
& CTC $\rightarrow$ +LLM & \multicolumn{2}{|c|}{CTC $\;/\;$ W{\tiny (ith)}B{\tiny (lank)} $\;/\;$ N{\tiny (o)}B{\tiny (lank)}} & WB $\;/\;$ NB & \multicolumn{1}{|c}{CTC + LLM} \\
\cline{2-6}
&\multicolumn{2}{c|}{Libri-other {\tiny (in-domain)}} & \multicolumn{1}{c|}{VP {\tiny (out-of-domain))}} & \multicolumn{2}{c}{CoVoST2 {\tiny (en-de)}} \\

\midrule
\multicolumn{6}{c}{\textbf{Stage~2}} \\
\hline

Ours & 4.60 $\rightarrow$ 4.57 &  {\tiny -} / 4.28 / 5.83 &  15.95 / 13.86 / 15.09 & 5.80 / \underline{19.97} {\tiny (49.4/26.1/15.1/9.1)} & 19.88 {\tiny (48.8/25.7/14.8/8.9)} \\

\hline

Ours\_TBP & \underline{4.38} $\rightarrow$ \underline{4.38} &  {\tiny -} / \underline{4.06} / 4.43 & 15.74 / 13.89 / 14.42 &  11.44 / 19.79 {\tiny (49.0/25.8/15.0/9.0)} &  \underline{20.10} {\tiny (49.3/26.2/15.2/9.1)} \\

\hline
\end{tabular}
}
\end{table}

\clearpage
\subsection{Pseudocode}
\label{sec:appendix_algorithms}

Algorithm \ref{algorithm:1} provides the full training procedure for DirectSpeech2LLM. The key steps are: (i) computing distance-based logits over the frozen LLM embedding matrix, (ii) deriving greedy CTC labels for guided downsampling and commitment loss, and (iii) optionally applying LLM cross-entropy in Stage 2. 

Algorithm \ref{algorithm:ctc_vocab} describes the construction of the speech tokenizer vocabulary $\mathcal{V}_{\text{ctc}}$ as the intersection of a BPE vocabulary trained on ASR transcripts and the LLM vocabulary.

Note that the commitment loss is clamped to never exceed the CTC loss. Because commitment targets are derived from greedy CTC predictions, a poorly calibrated CTC head can produce noisy labels, which would push SFM embeddings toward incorrect geometric anchors. By clamping the commitment loss, we ensure that it cannot dominate training before CTC has converged sufficiently.

\begin{algorithm}[ht]
\caption{DirectSpeech2LLM method}
\label{algorithm:1}
\begin{algorithmic}[1]

\Require Training set $\mathcal{D}=\{(X^{(i)},Y^{(i)})\}_{i=1}^N$
\Require SFM $E_{\phi}$ with output dimension $d_{\text{llm}}$
\Require LLM $\mathcal{M}_\theta$ and its embedding matrix $Q\in\mathbb{R}^{|\mathcal{V}_{\text{llm}}|\times d_{\text{llm}}}$ 
\Require Speech tokenizer vocab $\mathcal{V}_{\text{ctc}}\subset\mathcal{V}_{\text{llm}}$
\Require Blank index $b$
\Require Loss weights $\lambda_{\text{llm}}$

\Statex
\For{each training iteration}

    \State $S \gets E_{\phi}(X)$
        \Comment{$S\in\mathbb{R}^{B\times T\times d_{\text{llm}}}$}

    \State Compute logits, $Z$, using 
    
    \[ 
       Z_{t,v}
        =
        -\left(
        \lVert s_t \rVert_2^2
        - 2 s_t^\top q_v
        + \lVert q_v \rVert_2^2
        \right)
    \]
    
     \State $P \gets \texttt{Log-Softmax}(Z)$
        \Comment{$P \in\mathbb{R}^{B\times T\times \mathcal{V}_{\text{llm}}}$}
     \State $P_{\text{ctc}} \gets P[:,:, \mathcal{V}_{\text{ctc}}]$
        \Comment{Gather valid $\mathcal{V}_{\text{ctc}}$ log-prob}
     \State $\mathcal{L}_{\text{ctc}} \gets \texttt{CTC\_Loss}(P_{\text{ctc}}), Y)$

    \Statex
    \State $\hat{I} \gets \arg\max_{v \in \mathcal{V}_{\text{ctc}}} P_{\text{ctc}}$
            \Comment{Greedy token indices}
    \State $S_q \gets Q[\hat{I}]$
        \Comment{Quantized embeddings}
    \State $M \gets \mathbbm{1}[\hat{I}_t \neq b]$
        \Comment{Non blank mask}
    \State $\mathcal{L}_{\text{commit}} \gets
        \dfrac{\lVert (S - S_q)\odot M \rVert_F^2}{\sqrt{d_{\text{llm}}}}$
        
    \Statex 
    \State $S_{\text{seg}} \gets
    \texttt{CTC-Guided-Downsampling
}(S, \hat{I})$
    
    \Statex 
    \State $E_p \gets Q(\texttt{Prompt})$ 
    \Comment{Prompt embeddings}
    \State $\mathcal{L}_{\text{llm}} \gets
    \texttt{CrossEntropy}(
        \mathcal{M}_{\theta}(E_p, S_{\text{seg}}), Y)$

    \Statex 
    \State $\mathcal{L} \gets
        \mathcal{L}_{\text{ctc}}
        + \mathcal{L}_{\text{commit}}
        + (\mathcal{L}_{\text{llm}}\times\lambda_{\text{llm}})$

    \State Update $\phi$ using $\nabla \mathcal{L}$

\EndFor

\end{algorithmic}
\end{algorithm}

\begin{algorithm}[ht]
\caption{Speech tokenizer vocabulary}
\label{algorithm:ctc_vocab}
\begin{algorithmic}[1]

\Require Text corpus $\mathcal{C}$, target vocab size $K$
\Require LLM tokenizer with vocabulary $\mathcal{V}_{\text{llm}}$

\State Train a BPE tokenizer on $\mathcal{C}$ with size $K$
\State Obtain vocabulary $\mathcal{V}_{\text{bpe}}$ and merges $\mathcal{M}_{\text{bpe}}$

\State $\mathcal{V}_{\text{ctc}} \gets \mathcal{V}_{\text{bpe}} \cap \mathcal{V}_{\text{llm}}$

\State $\mathcal{M}_{\text{ctc}} \gets \{(a,b)\in\mathcal{M}_{\text{bpe}} \mid a,b,(a\Vert b)\in\mathcal{V}_{\text{ctc}}\}$

\State Construct speech tokenizer using $\mathcal{V}_{\text{ctc}}$ and $\mathcal{M}_{\text{ctc}}$

\State \Return speech tokenizer defining $\mathcal{V}_{\text{ctc}}\subset\mathcal{V}_{\text{llm}}$

\end{algorithmic}
\end{algorithm}

\subsection{Task prompts}
\label{sec:appendix_taskpromptsused}

The template used for \texttt{Qwen-3} LLM is shown in Figure \ref{fig:taskprompt}.

\begin{figure}[h]
\centering
\begin{mybox}{-}
\noindent\makebox[\linewidth]{\hrulefill\ \textbf{Prompt Template}\ \hrulefill}

\begin{minipage}{\linewidth}
\vspace{4pt}
\ttfamily\small
\textless|im\_start|\textgreater system\par
You are a helpful Multilingual conversational assistant.\par
\textless|im\_end|\textgreater\par
\textless|im\_start|\textgreater user\par
\underline{TASK SPECIFIC PROMPT}\par
\textless context\textgreater\par
\underline{SFM OUTPUT EMBEDDINGS}\par
\textless/context\textgreater\par
\textless|im\_end|\textgreater\par
\textless|im\_start|\textgreater assistant\par
\textless think\textgreater\textbackslash n\par
\textless/think\textgreater\textbackslash n\par
\dots \underline{LLM RESPONSE} \dots
\vspace{12pt}
\end{minipage}

\noindent\makebox[\linewidth]{\hrulefill\ \textbf{Task Specific Prompts}\ \hrulefill}

\begin{minipage}{\linewidth}
\vspace{4pt}
\ttfamily\small
\textbf{ASR: }
Based on the context, correctly repeat the text without any explanations or formatting.\\
\textbf{ST: }
Based on the context, correctly translate the text into German/Chinese. Output only the German/Chinese translation without any explanations or formatting.\\
\textbf{ER: }
Based on the context, identify the emotion from the following categories: angry, happy, neutral, and sad. Output only the emotion category without any explanations or formatting.
\end{minipage}

\end{mybox}
\caption{Prompt template used for Stage 2 training and inference with Qwen3, including task-specific prompts for ASR, speech translation, and emotion recognition.}
\label{fig:taskprompt}

\end{figure}

\clearpage
\subsection{Discussion}
\label{sec:appendix:discussion_limitation}
Failure examples from Stage~1 using no-blank (NB) decoding for the ASR task are shown in Table \ref{tab:errorsstages}. These failures are due to the input distribution mismatch between the CTC and LLM tokenizers or sometimes the LLM ignores the instruction and starts translating. These errors are eliminated when using the same decoding strategy as used during training, WB or NB.

One potential direction to address the formatting errors is to use byte-level LLMs \cite{xue2022byt5, yu2023megabyte}. Although byte-level modeling increases sequence length, recent work such as the Byte Latent Transformer (BLT) \cite{pagnoni2025blt} addresses this by learning latent token representations that compress byte sequences dynamically while preserving modeling capacity providing a promising direction for bridging CTC style character level embeddings with language models. Similar to \cite{tsiamas2024pushingspeechtransaltionctclaignnmentoptimaltransport2}, we propose a simple solution in Section~\ref{sec:appendix_hack}, called TBP, to make any LLM take raw-bytes/characters as input and therefore extending it to  DirectSpeech2LLM. The results are shown in Table~\ref{tab:ext_main_results} with marginal improvement on the ASR task and no clear gains on the unseen tasks. 

Our method's sensitivity to the duration of input speech seen in training vs inference is shown in \ref{sec:appendix_longform}. During training it sees speech upto 20 seconds and is able to do inference upto 300 seconds after which the performance lags behind cascaded system.

Lastly, it is easier to align spoken text temporally with a LLM. However, it is non-trivial to align audio with a LLM which do not have any direct temporal relation to align, especially if the LLM was never exposed to audio during pre-training.
Expecting such a model to suddenly perform multimodal instruction following is unrealistic. It is analogous to bypassing the foundational text pre-training phase in LLM development, without learning the basic structure and representations of the modality, consistent performance is not achievable.

\begin{table}[ht]
\caption{Dominant failure cases of DirectSpeech2LLM on the ASR task when using the NB decoding strategy. Stage~1 formatting failures are expected as we ask the LLM to repeat. Even after stage~2 LLM finetuning, the formatting failures of limited vocabulary of speech tokenizer are still visible, very rare. We also found degradation in instruction following as sometimes, instead of repeating, the llm starts translating.}
\label{tab:errorsstages}
\centering
\begin{tabular}{p{\columnwidth}}

\toprule
\textbf{Stage~1} -- Failure: Formatting \\
\hline
\textbf{Real:} i'd recommend him to you instead of blackstone thanks laughed kenneth \\
\textbf{CTC:} i'd recommend him to you instead of blackstone thanks laupped kenneth \\
\textbf{LLM (NB):} i d e r c o m e n d h i m t o y o u i n s t e a d o f b a c k s t o n e t h a n s l a l p e d k e n n e t h \\

\midrule
\midrule

\textbf{Stage~2} -- Failure: Formatting / Translation \\
\hline
\textbf{Real:} if they tried to run they were hit from behind if they stood still they were clubbed carefully \\
\textbf{CTC:} if they tried to run they were hit from behind if they stood still they were clubbed carefully \\
\textbf{LLM (WB):} if they tried to run they were hit from behind if they stood still they were clubbed carefully \\
\textbf{LLM (NB):} iftheytriedtoruntheywerehitfrombehindiftheystoodstilltheywereclubbedcarefully \\
\hline

\textbf{Real:} oh papa by that testing everything by the standard of wealth \\
\textbf{CTC:} oh papa by that testing everything by the standard of wealth \\
\textbf{LLM (WB):} oh papa by that testing everything by the standard of wealth \\
\textbf{LLM (NB):} o papá porque tudo é avaliado pelo padrão da riqueza \\

\bottomrule
\end{tabular}
\end{table}

\subsection{Effect of Restricted CTC Vocabulary as input to LLM}

When SFM outputs are tokenized using the restricted vocabulary $V_{\text{ctc}}$ and fed directly to the LLM (without Stage 2 fine-tuning), the LLM generates outputs at a different granularity than expected, most commonly producing character-level or sub-word-level output rather than the word level transcriptions seen during the LLM's own pretraining. Table \ref{tab:llminputablations} quantifies this effect by measuring WER when ground-truth transcripts (rather than speech) are used as input, thereby isolating the impact of tokenizer mismatch from acoustic modeling errors.
The results reveal two notable patterns. First, Phi-3.1-3.8B is considerably more sensitive to vocabulary mismatch than Qwen3-4B. Second, no clear scaling trend is observed across Qwen3 model sizes, suggesting that the issue is not simply a function of model capacity but rather of how well each LLM accommodates restricted vocabulary tokenization.
Importantly, the elevated WER values reported under $V_{\text{ctc}}$ in Table \ref{tab:llminputablations} are attributable primarily to formatting errors (character-level outputs, missing spaces, occasional translation) rather than genuine transcription errors. These errors are resolved after Stage 2 finetuning, as shown in the main results.

\begin{table}[ht]
\centering
\caption{Effect of using a $\sim$1k subset of the LLM vocabulary for tokenizing SFM-OES. When evaluated on LibriSpeech \textit{test-other} using ground-truth transcripts as input to the ASR prompt. Performance degrades substantially under $V_{\text{ctc}}$, with Phi-3.1-3.8B showing higher sensitivity than Qwen-3-4B. No clear scaling trend is observed across Qwen models. The reader should keep in mind that inflated WER is because of formatting errors as shown in Table~\ref{tab:errorsstages} in the Appendix. Furthermore, these errors to a large extent are resolved after finetuning.}
\label{tab:llminputablations}
\begin{tabular}{lcc}
\toprule
\textbf{Model} & \textbf{Vocab} & \textbf{WER (\%)} \\

\hline
\hline
\multicolumn{3}{c}{\textit{Base comparison}} \\
\midrule
Phi-3.1-3.8B & \multirow{2}{*}{$V_{\text{llm}}$} & 0.91 \\
Qwen-3-4B    &  & 0.02 \\
\cline{2-2}
Phi-3.1-3.8B & \multirow{2}{*}{$V_{\text{ctc}}$} & 6.91 \\
Qwen-3-4B    & & 4.40 \\

\hline
\hline
\multicolumn{3}{c}{\textit{Qwen scaling under $V_{\text{ctc}}$}} \\
\hline
Qwen-3-0.6B  & \multirow{5}{*}{$V_{\text{ctc}}$} & 6.46 \\
Qwen-3-1.7B  & & 4.89 \\
Qwen-3-8B    & & 6.33 \\
Qwen-3-14B   & & 3.03 \\
Qwen-3-32B   & & 10.37 \\

\bottomrule
\end{tabular}
\end{table}


\end{document}